\documentclass{article}

\usepackage{amsmath,amsfonts,bm}
\usepackage{graphicx} %
\usepackage{amsmath}
\usepackage{amssymb}
\usepackage{natbib}
\usepackage{booktabs}

\usepackage{amsthm}
\usepackage[most]{tcolorbox}
\usepackage{algorithm}
\usepackage{appendix}

\def\eqref#1{equation~\ref{#1}}

\def\1{\bm{1}}

\usepackage{xparse}
\usepackage{xspace}
\usepackage{algorithm}%
\usepackage{fixltx2e}
\usepackage{tikz}
\usepackage{float}
\usepackage{multirow}
\usepackage{multicol}
\usepackage{colortbl}
\usepackage{array}
\newcommand{\PreserveBackslash}[1]{\let\temp=\\#1\let\\=\temp}
\newcolumntype{C}[1]{>{\PreserveBackslash\centering}p{#1}}
\newcolumntype{R}[1]{>{\PreserveBackslash\raggedleft}p{#1}}
\newcolumntype{L}[1]{>{\PreserveBackslash\raggedright}p{#1}}

\usepackage{microtype}
\usepackage{graphicx}
\usepackage{mathtools}
\usepackage{float}
\usepackage{subcaption}
\usepackage{booktabs} %
\usepackage[shortlabels]{enumitem}

\usepackage{amssymb}%
\usepackage{pifont}%

\usepackage{bbold}

\usepackage{hyperref,amssymb,enumitem}

\setlist[itemize]{leftmargin=*}
\setlist[enumerate]{leftmargin=*}

\newcommand{\mycolspace}{0pt}

\newcommand{\cmark}{\ding{51}}%
\newcommand{\xmark}{\ding{53}}%

\newcommand*{\rej}{{\ooalign{\lower.3ex\hbox{$\sqcup$}\cr\raise.4ex\hbox{$\sqcap$}}}}

\usepackage{stackengine}

\usepackage{xcolor}

\newcommand{\ie}{\textit{i.e.,}\@\xspace}
\newcommand{\eg}{\textit{e.g.,}\@\xspace}

\newcommand{\validation}{held-out\@\xspace}
\newcommand{\Validation}{Held-out\@\xspace}

\newcommand{\encircle}[1]{%
  \tikz[baseline=(X.base)] 
    \node (X) [draw, shape=circle, inner sep=0pt] {\strut #1};%
}

\newcommand{\hnull}{\mathcal{H}_{0}}

\newcommand{\dtarget}{\mathcal{D}_{\text{sus}}}
\newcommand{\dsus}{\mathcal{D}_{\text{sus}}}
\newcommand{\dval}{\mathcal{D}_{\text{val}}}
\newcommand{\dtextdiff}{\mathcal{D}_{\text{text}}^{\text{diff}}}
\newcommand{\dcombdiff}{\mathcal{D}_{\text{comb}}^{\text{diff}}}
\newcommand{\dtargettrain}{\mathcal{D}_{\text{sus}}^{\text{train}}}
\newcommand{\dsustrain}{\mathcal{D}_{\text{sus}}^{\text{train}}}
\newcommand{\dvaltrain}{\mathcal{D}_{\text{\,val}}^{\text{\,train}}}
\newcommand{\dtargettest}{\mathcal{D}_{\text{\,sus}}^{\text{\,test}}}
\newcommand{\dsustest}{\mathcal{D}_{\text{\,sus}}^{\text{\,test}}}
\newcommand{\dvaltest}{\mathcal{D}_{\text{\,val}}^{\text{\,test}}}

\newcommand{\xtarget}{x_{\text{sus}}}
\newcommand{\xsus}{x_{\text{sus}}}
\newcommand{\xval}{x_{\text{val}}}

\newcommand{\xsustest}{x_{\text{sus}}^{\text{test}}}
\newcommand{\xvaltest}{x_{\text{val}}^{\text{test}}}

\newcommand{\auctext}{\textrm{AUC}_{\textrm{ Text}}}
\newcommand{\auccomb}{\textrm{AUC}_{\textrm{ Comb}}}
\newcommand{\ctext}{c_{\text{text}}}
\newcommand{\ccomb}{c_{\text{comb}}}
\newcommand{\mia}{\text{MIA}}
\newcommand{\expect}{\mathds{E}}
\newcommand{\real}{\mathds{R}}
\newcommand{\ydiff}{y_{\text{diff}}}

\graphicspath{{images/}}

\usepackage{booktabs,arydshln}
\makeatletter
\def\adl@drawiv#1#2#3{%
        \hskip.5\tabcolsep
        \xleaders#3{#2.5\@tempdimb #1{1}#2.5\@tempdimb}%
                #2\z@ plus1fil minus1fil\relax
        \hskip.5\tabcolsep}
\newcommand{\cdashlinelr}[1]{%
  \noalign{\vskip\aboverulesep
           \global\let\@dashdrawstore\adl@draw
           \global\let\adl@draw\adl@drawiv}
  \cdashline{#1}
  \noalign{\global\let\adl@draw\@dashdrawstore
           \vskip\belowrulesep}}
\makeatother

\usepackage[capitalize,noabbrev]{cleveref}

\crefalias{prop}{proposition} %

\newcommand{\nlp}[1]{}

\newcolumntype{x}[1]{>{\centering\arraybackslash\hspace{0pt}}p{#1}}

\usepackage[textsize=tiny]{todonotes}

\newcommand{\green}{\textcolor{teal}}
\newcommand{\red}{\textcolor{red}}

\definecolor{chocolate(traditional)}{rgb}{0.48, 0.25, 0.0}

\definecolor{darkpastelgreen}{rgb}{0.01, 0.75, 0.24}

\definecolor{pistachio}{rgb}{0.58, 0.77, 0.45}

\definecolor{amber(sae/ece)}{rgb}{1.0, 0.49, 0.0}

\usepackage{microtype}
\usepackage{graphicx}
\usepackage{booktabs} %
\usepackage{listings}

\usepackage{xurl}
\usepackage{hyperref}
\usepackage{algorithm}
\usepackage{algorithmic}

\usepackage[accepted]{icml2025}

\usepackage{amsmath}
\usepackage{amssymb}
\usepackage{mathtools}
\usepackage{xcolor}
\usepackage{dsfont}
\usepackage[flushleft]{threeparttable}

\usepackage[capitalize,noabbrev]{cleveref}

\crefname{algorithm}{algorithm}{algorithms}
\Crefname{algorithm}{Algorithm}{Algorithms}
\crefname{algocf}{algorithm}{algorithms}
\Crefname{algocf}{Algorithm}{Algorithms}
\crefalias{algocf}{algorithm}
\crefalias{algorithm}{algorithm}
\crefname{algorithm}{algorithm}{algorithms}
\Crefname{algorithm}{Algorithm}{Algorithms}

\crefalias{ALG@line}{algorithm}
\crefalias{algocf}{algorithm}
\crefalias{algorithmcf}{algorithm}

\theoremstyle{plain}

\theoremstyle{definition}

\theoremstyle{remark}

\usepackage[textsize=tiny]{todonotes}

\newcommand{\ourtitle}{Unlocking Post-hoc Dataset Inference with Synthetic Data} %

\icmltitlerunning{\ourtitle}

\begin{document}

\twocolumn[
\icmltitle{\ourtitle}

\icmlsetsymbol{equal}{*}

\begin{icmlauthorlist}
\icmlauthor{Bihe Zhao}{cispa}
\icmlauthor{Pratyush Maini}{cmu,dat}
\icmlauthor{Franziska Boenisch}{cispa}
\icmlauthor{Adam Dziedzic}{cispa}
\end{icmlauthorlist}

\icmlaffiliation{cispa}{CISPA Helmholtz Center for Information Security}
\icmlaffiliation{cmu}{Carnegie Mellon University}
\icmlaffiliation{dat}{DatologyAI}

\icmlcorrespondingauthor{Bihe Zhao}{bihe.zhao@cispa.de}
\icmlcorrespondingauthor{Pratyush Maini}{pratyushmaini@cmu.edu}
\icmlcorrespondingauthor{Franziska Boenisch}{boenisch@cispa.de}
\icmlcorrespondingauthor{Adam Dziedzic}{adam.dziedzic@cispa.de}

\icmlkeywords{Machine Learning, ICML}

\vskip 0.3in
]

\printAffiliationsAndNotice{}  %

\begin{abstract}

The remarkable capabilities of Large Language Models (LLMs) can be mainly attributed to their massive training datasets, which are often scraped from the internet without respecting data owners’ intellectual property rights. Dataset Inference (DI) offers a potential remedy by identifying whether a suspect dataset was used in training, thereby enabling data owners to verify unauthorized use. However, existing DI methods require a private set—known to be absent from training—that closely matches the compromised dataset’s distribution. Such in-distribution, held-out data is rarely available in practice, severely limiting the applicability of DI. In this work, we address this challenge by synthetically generating the required held-out set. Our approach tackles two key obstacles: (1) creating high-quality, diverse synthetic data that accurately reflects the original distribution, which we achieve via a data generator trained on a carefully designed suffix-based completion task, and (2) bridging likelihood gaps between real and synthetic data, which is realized through post-hoc calibration. Extensive experiments on diverse text datasets show that using our generated data as a held-out set enables DI to detect the original training sets with high confidence, while maintaining a low false positive rate. This result empowers copyright owners to make legitimate claims on data usage and demonstrates our method’s reliability for real-world litigations.
Our code is available at \url{https://github.com/sprintml/PostHocDatasetInference}.

\end{abstract}

\begin{figure*}[t]
	\centering	\includegraphics[width=0.85\textwidth]{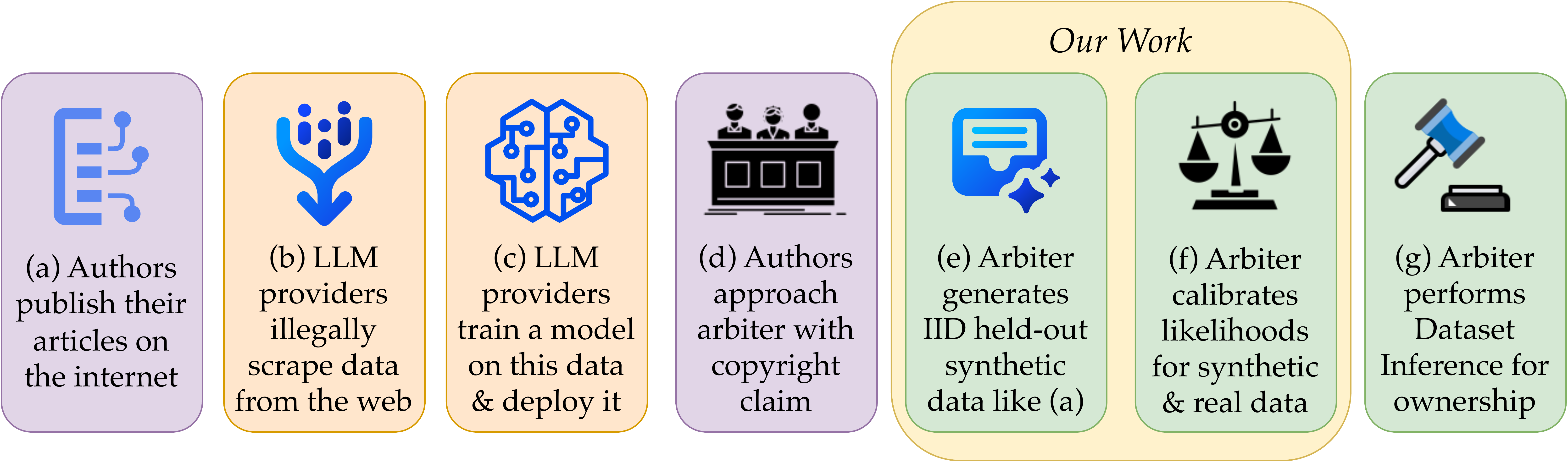}
    \vspace{-5pt}
	\caption{\textbf{Dataset Inference Procedure with Synthetic Held-Out Data.} 
    This figure presents a high-level overview of how the proposed dataset inference (DI) process will take place in real-world use cases.
    \textbf{(a-d)} LLM providers scrape proprietary author data from the internet, and train an LLM on it. Authors who suspect unauthorized use may approach an arbiter with a copyright claim.
    To resolve such a dispute, the arbiter must perform dataset inference \textbf{(g)}. However, this requires the presence of a held-out dataset that is IID to the suspect set. \textbf{Our Work:} 
    \textbf{(e)} The arbiter generates IID synthetic held-out data that mimics the author’s original data.
    \textbf{(f)} The arbiter calibrates likelihoods between real and synthetic data to ensure fair comparison, enabling them to reliably perform dataset inference.
    }
    \label{fig:teaser}
    \vspace{-5pt}
\end{figure*}

\section{Introduction}
\label{sec:intro}

Large language models (LLMs) have recently achieved remarkable success in a broad range of tasks, fueled by the availability of massive high-quality text corpora often scraped from the internet~\citep{weber2024redpajama,penedo2024fineweb}. While this practice enables LLMs to generate high-quality text and to excel on benchmarks, it also raises serious concerns related to intellectual property rights~\citep{gettyimages_lawsuit,Grynbaum2023LawsuiteOpenAINewYorkTimes,SilvermanLawsuitOpenAI}, data privacy~\citep{duan2023flocks,duan2023privacyICL,hanke2024openLLMs,hayes2025strong}, and transparency~\citep{rahman2023BeyondFairUse,wu2023unveiling}. The reliance on potentially unauthorized data creates an urgent need for methods that allow independent authors to verify whether a given dataset has been used to train an LLM without the explicit consent of the model provider.

A promising approach to addressing these concerns is \emph{dataset inference} (DI)~\citep{maini2021dataset,sslextractions2022icml, dziedzic2022dataset,maini2024llm,dubinski2024cdi,kowalczuk2025privacy}, which aims to determine whether a suspect dataset has contributed to a model's training. This puts power in the hands of data owners to monitor and exercise their intellectual property rights. 
Despite its potential, DI currently faces a critical bottleneck: it requires a \validation set---a dataset known to be absent from training---that shares the same distribution as the suspect dataset~\citep{zhang2024membership}. 
In practice, however, such an in-distribution \validation set is rarely available. Data creators do not typically reserve a dedicated \validation set for legal or auditing purposes, and any disclosed \validation data could itself be repurposed for future training, further complicating the verification process. Moreover, even when a dataset owner can provide some \validation samples, any slight distributional discrepancy from the original suspect data can undermine DI by inflating false positives~\citep{das2024blind,duan2024membership,meeus2024sok,maini2024reassessingemnlp2024s}. 

To illustrate the brittleness of using seemingly IID (Independent and Identically Distributed) \validation data, we demonstrate in \Cref{sec:fail_case} that even in a simple scenario—where an LLM is fine-tuned on blog posts from a \emph{single} author—there exists a distributional shift between training data (members) and randomly held-out blog posts from the same author. This highlights how even subtle variations in \validation data can undermine DI. Malicious actors may exploit this vulnerability by strategically introducing \emph{shifted} \validation data, falsely accusing model owners of copyright infringement and further reducing the reliability of DI methods.

In this work, we address these challenges by proposing to \textit{synthetically generate} \validation data for DI, bypassing the need for in-distribution held-out data. This vision, however, is non-trivial to achieve. First, the generated texts must be realistic, high-quality, and sufficiently diverse to approximate the distribution of the original data. Second, the generation process itself may introduce a distribution shift between natural and synthetic held-out data. Such a shift complicates DI: if a difference is observed between the suspect and \validation sets, it becomes unclear whether this difference arises from a genuine membership signal (\ie the target model behaves differently on the suspect data because it has seen it during training) or merely from the distribution shift (\ie the model behaves differently on suspect data because it is natural data). Recent studies have extensively highlighted this issue in the context of Membership Inference Attacks (MIAs)~\citep{shokri2017membership}, where distribution shifts lead to misleading evaluation results~\citep{das2024blind,zhang2024membership,maini2024llm,dubinski2024cdi}.

To this end, we first train a carefully designed text generator on the suspect dataset itself, on a suffix completion task (\Cref{sec:gen_method}). This approach produces high-quality datasets with only a small distributional shift from the suspect texts.
However, even small shifts in distribution can undermine DI’s reliability. To address this, we introduce a \emph{post-hoc} calibration step (\Cref{sec:calibration}) to ensure that the generated \validation set can serve as a reliable reference for DI. Specifically, we disentangle the effects of distributional shifts from the actual membership signal---a critical factor in DI. To achieve this, we propose a dual-classifier approach:  
(1) A \emph{text-only classifier}, trained to distinguish natural (original) from generated data.  
(2) A \emph{membership-aware classifier}, which incorporates both the textual features and DI’s standard membership indicators (e.g., perplexity, min-k probabilities).  
The key insight is that any performance advantage of the membership-aware classifier over the text-only classifier must arise from the presence of membership signals rather than distributional artifacts. This difference serves as our DI signal, allowing us to infer whether the suspect dataset was used in the target model’s training. This calibration strategy enhances DI’s robustness, reducing false positives while maintaining high detection accuracy.

We demonstrate the effectiveness of our approach on diverse textual datasets, ranging from single-author datasets (e.g., personal blog posts) to large-scale, multi-author collections such as Wikipedia. Our results show that using \emph{synthetic} \validation data, combined with calibration, enables DI to detect unauthorized training data use with high confidence while keeping false positives low. This expands the practical applicability of DI and provides a pathway for data owners to safeguard their intellectual property in an era of LLMs.

\section{Background and Related Work}

\subsection{Membership Inference}
MIAs focus on deciding if a single data point was included in a given model's training dataset and often serve as features extractors for DI.
In the LLM domain, MIAs exploit different signals to distinguish between members (training data points) and non-members (data points not used during training).
For instance, LOSS exploits the perplexity or loss function of the target model \citep{yeom2018privacy}. 
\citet{shi2024detecting} find that the rare words in a sequence can leak more privacy information, and select K\% tokens with the smallest probabilities for evaluation. 
Min-K\%++ further improves upon the Min-K\% approach by introducing two calibration factors \citep{zhang2024min}.
Zlib ratio \citep{carlini2021extracting} uses the compression rate of z-library to normalize the perplexity of the target model.
Neighborhood-based methods compare a suspect sequence with its neighboring texts, which can be produced by synonym substitution \citep{mattern2023membership} or paraphrasing \citep{duarte2024cop}.
Moreover, reference-based methods compare the output signals on a suspect sample between the target model and a reference model  \citep{fu2024membership}.
Yet, many recent works have shown that the evaluation of MIAs suffers from a falsified experimental setup, where a distributional shift exists between the member and non-member sets \citep{ zhang2024membership, maini2024llm,das2024blind}.
\citet{duan2024membership} show that most MIAs only perform slightly better than random guessing if evaluated correctly on non-biased benchmarks.
Recently, \citet{kazmi2024panoramia} proposed how to de-bias MIAs from this distribution shift---which we use as a foundation for our DI calibration.

\subsection{Dataset Inference}
To strengthen the signal from training data further beyond MIAs, \citet{maini2021dataset} introduced DI. 
DI aggregates the membership signal over multiple data points, often referred to as \textit{suspect set}, to decide whether a given model was trained on this data.
More formally, given a target model $f$, DI aims to detect whether $f$ was trained on the suspect dataset $\dsus$.
Therefore, it needs an additional \validation dataset $\dval$ from the same distribution as $\dsus$.
Given both sets, DI extracts membership features from the data points in $\dsus$ and $\dval$, aggregates all features per given sample, and then scores these aggregate features through a scoring model.
The scores should be lower for members than for non-members. Then, DI performs statistical hypothesis testing on the scores of $\dsus$ and $\dval$.
The null hypothesis is that the average scores for $\dsus$ are higher than or equal to the scores for $\dval$. If the statistical test manages to reject this null hypothesis, this is a confident indicator that the data points from $\dsus$ are indeed members of model $f$'s training data. Otherwise, the test is considered inconclusive.

How to extract the best membership features from the data points varies based on the learning paradigm. For example, the original DI for supervised classification models~\citep{maini2021dataset}
designs a random walk strategy to estimate the distance between data points and the decision boundary of a supervised model. 
For self-supervised models, \citet{dziedzic2022dataset} use Gaussian Mixture Model to estimate the representational differences between the training dataset (members) and the test data.
Recent dataset inference methods for LLMs~\citep{maini2024llm}, Diffusion Models~\citep{dubinski2024cdi}, and Image Autoregressive Models~\citep{kowalczuk2025privacy} build on existing membership inference attacks (MIAs) tailored to each type of generative model. These methods extract membership-related features using the appropriate MIA and then apply a linear model to combine and weight the extracted features.
We follow this approach in our evaluations.
LLM DI can be formalized as follows.
First, after calculating over $n$ MIA scores with linear regression, an aggregated MIA score is obtained by $W\cdot \mia(x) = \sum_{i=1}^n w_i\mia_i(x)$.
Here, $W=[w_1,...,w_n]$ is the weight of the linear regressor, and $\mia(x)$ is a vector concatenating $n$ MIA scores.
We label the suspect data as 0 and the \validation data as 1.
Note that, $\mia(x)$ is calculated based on $f(x)$, but we omit $f$ for simplicity.
Then, a hypothesis testing is conducted to verify if the \validation set has higher MIA score than the suspect set statistically. 
The null hypothesis can be formalized as follows.
\begin{equation}
\begin{aligned}
    \hnull: \expect_{\dval}[W\cdot\mia(\xval)] \leq \expect_{\dsus}[W\cdot\mia(\xsus)].
\end{aligned}
\end{equation}
If the suspect set is part of the training set of $f$, the null hypothesis is rejected.

\section{Failure Cases of DI} \label{sec:fail_case}
In this section, we dive deeper into the difficulties that arise from DI's assumption on the availability of an additional in-distribution \validation dataset.
More precisely, we show that this assumption is extremely hard to meet in practice, even in the simplest setups---limiting the applicability of standard DI.
Therefore, we collect blog posts written by a \textit{single author} on topics from the \textit{same domain} and split them randomly into a training and \validation set.
We finetune an LLM on the training set, perform DI~\citep{maini2024llm}, and find that the method returns false positives, \ie it illegitimately claims that the model was trained on blog posts that it actually was not trained on (see \Cref{table:auc_blog}). Our analysis highlights that despite the texts' homogeneity, there is a small distributional shift between the suspect and \validation sets that is not even easily distinguishable by Blind Baselines \cite{das2024blind},  which causes DI to fail.
This highlights the need to generate synthetic \validation data to benefit from DI in real-world copyright claims.
We provide more details below and discuss its implications.

\begin{table}[t!]
    \scriptsize
    \caption{\textbf{The distributional shift (GPT2 AUC) and DI p-value between a suspect set that consists of \textit{non-members} and \validation blog posts.} Here, p-value $<$ 0.05 indicates DI incorrectly suggests that the suspect set is a member set.}
    \begin{center}
    \begin{tabular}{cccccc}
        \toprule
        Sequences per Blog & 5 & 10 & 15 & 20 & 25 \\
        \cmidrule{1-6}\morecmidrules\cmidrule{1-6}
        GPT2 AUC (\%) & 52.0 & 55.2 & 53.2 & 58.2 & 58.6 \\
        DI p-value & 0.002 & $<$0.001 & $<$0.001 & $<$0.001 & $<$0.001 \\
        True Membership & \green{\xmark} & \green{\xmark} & \green{\xmark} & \green{\xmark} & \green{\xmark} \\
        Inferred Membership & \red{\cmark} & \red{\cmark} & \red{\cmark} & \red{\cmark} & \red{\cmark} \\
        \bottomrule
    \end{tabular}
    \end{center}
    \label{table:auc_blog}
    \vspace{-12pt}
\end{table}

\subsection{DI on a Single Author's Data}
We consider a practical application of DI in copyright protection as detailed in \Cref{fig:teaser}.
In this scenario, an author has some published texts on the internet of which they believe that they were illegitimately used by an LLM provider to train their model. The author provides these published works to an arbitrator, as a suspect set and some non-published blog-posts as \validation set from the same distribution, \ie with the same style, topics, etc. 
Then, the arbitrator performs DI to resolve the copyright claims.

To evaluate this setup in practice, we collect blog posts of a public blogger.
The blogs are split into member, non-member, and \validation sets.
To avoid any potential temporal or topic distributional shifts, we randomly shuffle all the collected blogs before splitting.
In lack of the computational capacities to train an LLM from scratch, we finetune a Pythia model \citep{biderman2023pythia} on the member set.
The Pythia model is trained on the Pile dataset \cite{gao2020pile}, so we only used blogs after the release date of the Pile to ensure that none of the blogs is part of the pre-training data.
Also, we only finetune the target model on the member set for one epoch.
This is to evaluate the performance of DI and our method in the most strict scenario, as \citet{duan2024membership} show that MIAs perform better with more training epochs.
Finally, we run DI. %
 More detailed experiment configurations can be found in \cref{sec:exp_setting}.

\subsection{Metrics of Distributional Gap} \label{sec:dist_metric}
Before analyzing the results, we introduce the metrics we use to quantify the distributional shift between the suspect and \validation sets.
Following the approach of Blind Baselines \cite{das2024blind}, we formulate the measurement of the distribution gap between two text datasets as a classification problem.
In particular, the suspect set $\dsus$ is randomly split into a classifier training split $\dsustrain$ and a test split $\dsustest$.
The \validation set $\dval$ is also split into $\dvaltrain$ and $\dvaltest$ in the same vein.
Then, a classifier $g$ is optimized to distinguish the training splits $\dsustrain$ and $\dvaltrain$.
Finally, we calculate the area under the curve (AUC) score of the classifier on the test splits $\dsustest$ and $\dvaltest$, which is used to measure the distributional gap between $\dsus$ and $\dval$.

The design of the classifier decides how the texts are vectorized and if the discrepancies between texts can be sufficiently captured.
\citet{das2024blind} apply a bag-of-words (BoW) classifier, which can only detect the differences in terms of word frequency.
Instead, we build a GPT2-based classifier with two transformer blocks to also find the differences in grammar, content, styles, etc. between two text distributions.
We train the classifier from scratch to avoid the impact of any pre-training data.
Using only two transformer blocks of the GPT2 architecture avoids overfitting.

\subsection{False Positive of DI}
\begin{figure}[t] 
	\centering	\includegraphics[scale=0.41]{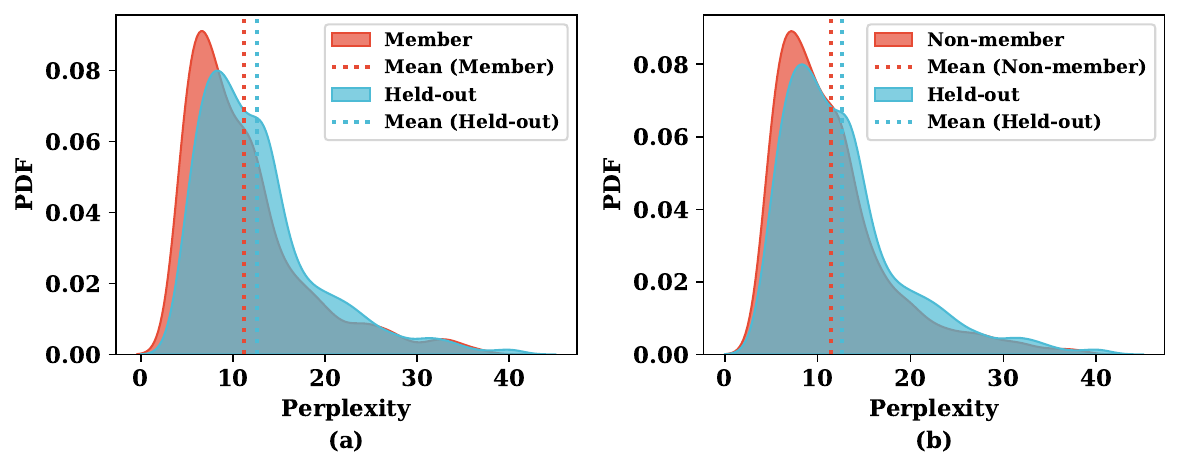}
    \vspace{-1em}
	\caption{
    \textbf{Probability Distribution Function (PDF) of target model perplexities.} We show the comparison between (a) the member and \validation, and (b) the non-member and \validation sets. 
    }
    \vspace{-1em}
    \label{fig:blog_dist}
\end{figure}

The AUC scores of the GPT2-based classifier in \cref{table:auc_blog} show that
there is a non-negligible distributional shift between the non-member and the \validation sets.
The intuition behind this observation is that each blog has different content and topics, which brings different words across the non-member and \validation documents.
The gap is enlarged when we sample more sequences from each blog post.
Consequently, this distributional shift in texts also lead to a shift in the MIA score.
As presented in \cref{fig:blog_dist}, the distributional shift in perplexities exists not only between member and \validation sets, but also between \textit{non-member and \validation sets}.
This shows that the inherent distributional shift among documents is entangled with the shift caused by membership signals in the MIA score, and makes DI fail to determine membership by simply detecting any distributional shift in the MIA score.
This observation aligns with the p-values and predictions in \cref{table:auc_blog}, where we find that even this small distributional shift causes significant false positive rates during DI.
This means that the DI falsely accuses the LLM provider of violating the copyright of an author.
What is more is that this shortcoming of DI can be maliciously exploited: authors could deliberately provide \validation data from a different distribution than their suspect data to mislead DI and \textit{illegitimately} accuse the LLM provider.
As a solution to this problem, in the next section, we propose our approach on generating an adequate in-distribution \validation dataset synthetically.

\begin{figure*}[t]
	\centering	\includegraphics[width=0.91\textwidth]{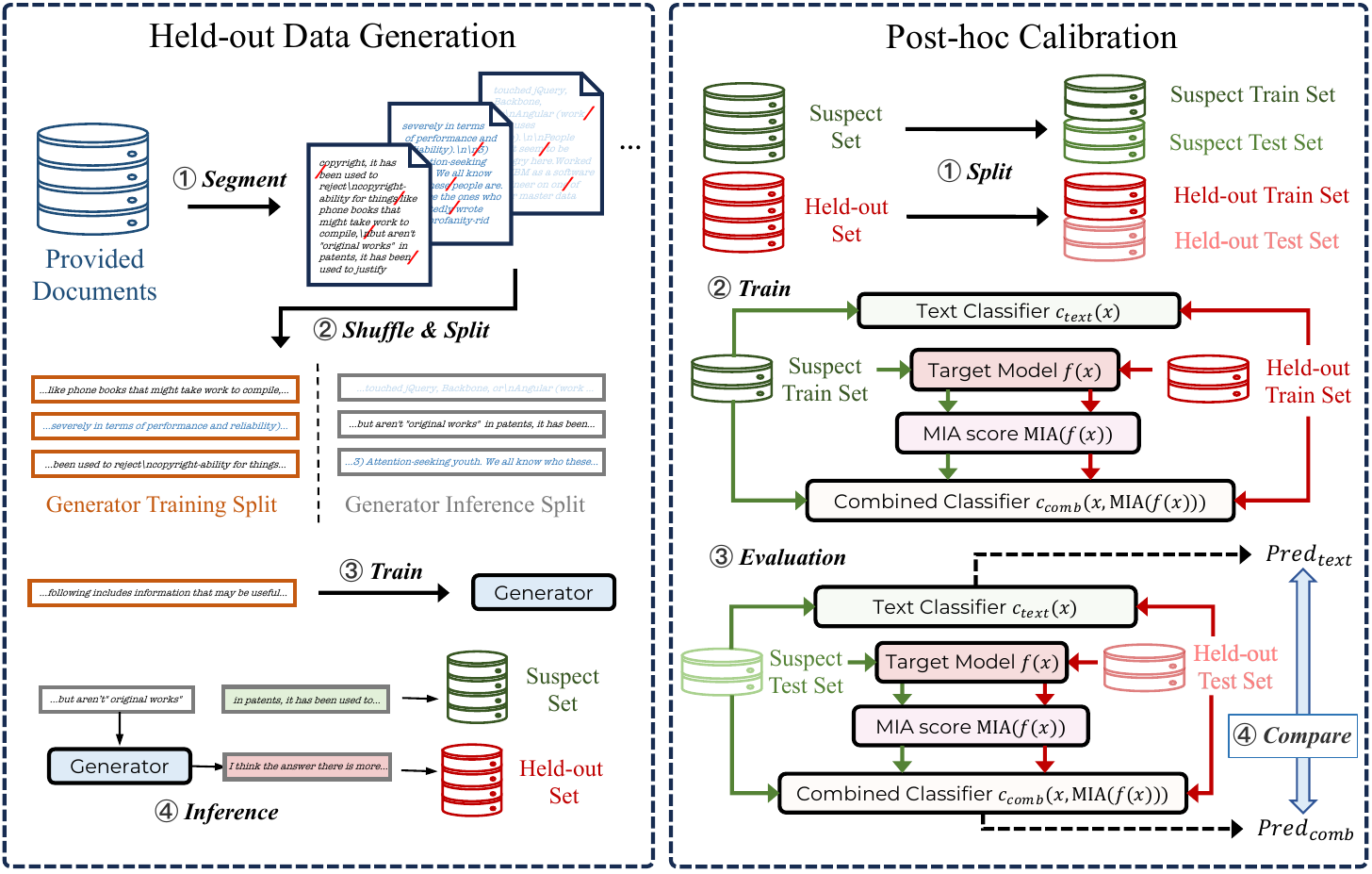}
        \vspace{-2pt}
	\caption{ 
    \textbf{Held-out Data Generation (Left Panel):} 
    \textbf{(1)} The suspect dataset is first segmented into text snippets.  
    \textbf{(2)} These snippets are shuffled and split into a generator training set and an inference set.  
    \textbf{(3)} A generator model is trained on the suspect dataset using a suffix completion task.  
    \textbf{(4)} The trained generator produces synthetic held-out data that mimics the suspect set.  
    \textbf{Post-hoc Calibration (Right Panel):}  
    \textbf{(1)} The suspect set and synthetic held-out set are split into training and test subsets.  
    \textbf{(2)} A \textbf{text classifier} is trained to differentiate real from synthetic text. A \textbf{combined classifier} integrates textual features with DI membership signals from the target LLM.  
    \textbf{(3)} The two classifiers are evaluated: the combined classifier should only outperform the text classifier if the suspect dataset was used for training.  
    \textbf{(4)} A statistical comparison ensures that any signal detected is due to actual membership rather than distributional shifts.
    }
	\label{fig:main_figure}
        \vspace{-9pt}
\end{figure*}

\section{Synthesizing \Validation Data}
Our approach consists of two subsequent steps. First, we generate high-quality \validation data, then, we perform a calibration to account for the distribution shift that such generation can introduce.

\subsection{\Validation Data Generation} \label{sec:gen_method}
We explore three approaches that leverage LLMs for generating \validation data based on provided suspect data with minimal distribution shift. %

\textbf{In-context Learning.}
As a na\"ive approach, we use GPT-4 models to paraphrase the suspect set with in-context learning (ICL) and evaluate the distributional shift between the original suspect texts and the paraphrased texts.
Specifically, each prompt includes a few data points as demonstrations (shots) and requests the model to produce paraphrases for another data point in the suspect set.
We test two types of prompts: 1) \textit{\textbf{Paraphrasing}}, where we directly prompt the model to paraphrase a chosen data point, and 2) \textit{\textbf{Text Completion}}, where we prompt the model to complete a truncated data point. For text completion, the original suffix is used as the suspect data, and the generated suffix as the held-out data.
Our results in \cref{table:gen_result} show that there is a significant distribution shift between the original and paraphrased samples for both types of prompts.
Even a BoW classifier obtains a significant AUC of 76.2\% when distinguishing between the original suspect vs paraphrased text and the GPT2 classifier can achieve 99.0\% AUC.
The reason is that there are many words (such as "remarkable" and "moreover")  that appear much more frequently in the synthetic text than in the human-written text. 
Please refer to \cref{app:gpt4} for more detailed explanation of GPT-4-based generation and examples of generated texts. 

\begin{table}[t!]
    \scriptsize
    \caption{\textbf{Distributional shifts between the suspect set and synthetic \validation set}. The shifts are measured by Bag-of-Word (BoW) and GPT2 classifiers.}
    \begin{center}
    \begin{tabular}{ccc}
        \toprule
          Generation Method & BoW AUC (\%) & GPT2 AUC (\%) \\
        \cmidrule{1-3}\morecmidrules\cmidrule{1-3}
        ICL Paraphrasing & 76.2 & 99.0 \\
        ICL Text Completion & 79.2 & 99.2 \\
        Preference Optimization & 50.2 & 58.9 \\
        Suffix Completion & \textbf{50.0} & \textbf{52.2} \\
        \bottomrule
    \end{tabular}
    \end{center}
    \label{table:gen_result}
    \vspace{-15pt}
\end{table}

\textbf{Preference Optimization.}
We also adapt preference optimization methods \citep{rafailov2024direct,xu2024contrastive} to the task of \validation data generation by changing from human preference to natural text preference.
A more detailed explanation is presented in \cref{sec:pref_opt}.
The AUC of the BoW classifier is similar to random guessing, which means frequent words can be greatly reduced in generated texts by this approach.
However, the GPT2 classifier can still obtain an AUC of 58.9\%.
This shows that preference optimization still leaves distinguishable generation patterns that could be easily captured by a transformer-based classifier, limiting the data's usefulness as in-distribution \validation set.

\textbf{Suffix Completion.}
The failure of the above methods demonstrates the difficulty of producing high-quality \validation data with a small enough distributional gap to the suspect data.
To solve this problem, we design a generator training scheme that enables the generator to derive a suspect set from the author's provided documents, together with a \validation set from the same distribution as this suspect set.
As shown in \cref{fig:main_figure}, we \encircle{1} first segment the provided documents into multiple short sequences.
\encircle{2} All the sequences are shuffled and randomly split into a generator training split and a generator inference split.
Then, \encircle{3} a low-rank adaption (LoRA) generator is finetuned on the training split with the cross-entropy loss for next-token prediction.
Finally, \encircle{4} we segment each sequence in the generator inference split into two parts, and the generator predicts a synthetic suffix based on the prefix.
Here, the original suffixes are used as the suspect set, and the synthetic suffixes as the \validation set.
Note that, the training and inference sets are split on the shuffled text sequences rather than on the documents.
This is to ensure that the text snippets from the generator training and inference splits are from the same distribution, such that the generator can achieve better generalization from the training to the inference set.
Furthermore, we design a suffix completion task for generator inference. In this task, both the original suffix and the synthetic suffix share a common prefix. This approach ensures that the synthetic text maintains the same position within a sentence as its original counterpart, making the two suffixes directly comparable. 
Another important insight is that the generator can produce suffixes of higher quality when the sequence length is relatively short.
Therefore, we limit the length of the sequences to no longer than 64 tokens for a smaller distributional gap. 
The results in \cref{table:gen_result} show that our method achieves a significantly small distributional shift, and even GPT2-based classifier can only achieve an AUC as low as 52.2\%.
For examples of our generative approach, please refer to \cref{app:gen_example}.

\subsection{Post-hoc Calibration}
\label{sec:calibration}
Since the generation itself can introduce a distributional shift (natural vs generated) data, DI might yield false positives. This is because it would detect differences between suspect and \validation data also when they only differ in terms of distribution but not necessarily in membership.
Therefore, we need to identify and mitigate this distribution shift. %

To do so, we rely on an important observation: the generation shift between natural and synthesized data occurs in the textual space, while the shift caused by the potential membership of the suspect set exists in the target LLM's output space. 
This allows us to disentangle the two signals.
By relying on our GPT-based \textbf{text-classifier} from \cref{sec:dist_metric}, we can quantify the textual distribution shift caused by the generation.
We denote this classifier by $\ctext(x)$, where $x$ is the text input for which the classifier should decide if it is original or generated data.
Inspired by \citet{kazmi2024panoramia}, we also define a second \textbf{MIA-classifier} with input signals from both the texts and the outputs of the target model, such that we can quantify the combined effects of generation and the membership signal.
Concretely, we train a combined classifier $\ccomb(x, \mia(f(x)))$ with inputs from both text $x$ and the MIA signal $\mia(x)$ based on the outputs of $f$.
Here, $\mia(x)$ can also be a vector by concatenating multiple MIA scores.
We split both the suspect set and \validation sets into training and test splits.
The two classifiers are optimized on the suspect train split $\dtargettrain$ and the \validation train split $\dvaltrain$, and evaluated on the suspect test split $\dtargettest$ and the \validation test split $\dvaltest$.
By comparing the distributional shifts quantified by the MIA classifier and the shifts identified by the text classifier,
we can separate the membership signals from the distribution gap caused by generation.

We design a hypothesis test to statistically verify if the combined classifier quantifies a larger distributional shift between the suspect and held-out data than the text classifier, namely the \textit{difference comparison t-test}.
The t-test is conducted on the test splits $\dsustest$ and $\dvaltest$, but we abbreviate them as $\dsus$ and $\dval$ for simplicity.
During the t-test, we first sample a suspect data point $\xsus\in\dsus$ and pair it with its corresponding generated counterpart $\xval\in\dval$.
Note that, $\xsus$ and $\xval$ are original and generated suffixes, which are both continuations of a common prefix.
For every such original/\validation pair, we quantify the shift caused by \textit{generation} with the text classifier as  $\ctext(\xval)-\ctext(\xtarget)$. We also quantify the combined effects caused by \textit{generation and membership signal} with the combined classifier as $\ccomb(\xval)-\ccomb(\xtarget)$.
In particular, $\xval$ and $\xtarget$ are labeled as 1 and 0, respectively. $c(x)$ denotes the predicted probability of classifier $c$ on data point $x$, which ranges between 0 and 1.
If the membership signal is present, the combined effects will be stronger than the generation effect alone, and the predicted probability will be slightly more accurate for the combined classifier, i.e. $\ccomb(\xval)-\ccomb(\xtarget) > \ctext(\xval)-\ctext(\xtarget)$.
To this end, we formalize the following null hypothesis for our t-test:
\begin{equation}
    \begin{aligned}
    \hnull:\ 
    & \expect_{\xval \in \dval, \xtarget \in \dtarget}[\ccomb(\xval)-\ccomb(\xtarget)] \leq \\
    & \expect_{\xval \in \dval, \xtarget \in \dtarget}[\ctext(\xval)-\ctext(\xtarget)].
    \end{aligned}
\end{equation}
The difference comparison t-test is performed multiple times with different random seeds, and the p-values are aggregated with Sidac correction~\citep{vsidak1967rectangular}. 

By introducing a dual-classifier approach along with a statistical test, we can statistically distinguish distributional shifts caused by actual membership signals from those caused by generation. 
Further results in \cref{sec:ablation} show that this approach can prevent false positives in dataset inference effectively.

\begin{figure}[t] 
	\centering	\includegraphics[width=0.38\textwidth]{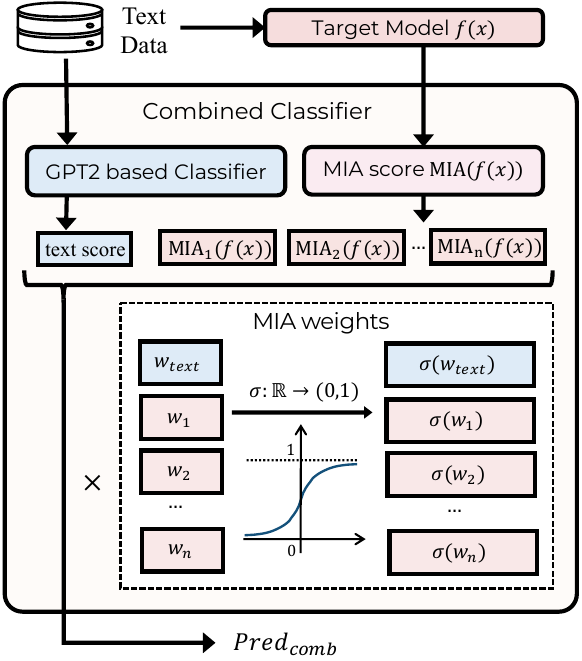}
    \vspace{-7pt}
	\caption{\textbf{Weight Constraint.}
     The weights for the MIA and text scores are constrained to (0,1) with Sigmoid function.
    }
    \vspace{-5pt}
    \label{fig:constraint}
\end{figure}

\subsection{Weight Constraint}\label{sec:constraint}
In this section, we explain why and how we apply a weight constraint when computing the importance of different MIA scores.
In the original DI, the aggregated MIA score is compared between the \validation and the suspect sets.
We define the difference in aggregated MIA score between the two sets $y_{\text{diff}}$ as follows:
\begin{equation}
\begin{aligned}
\ydiff \
&=\expect[\sum_{i=1}^n w_i\mia_i(\xval)] -\expect[\sum_{i=1}^n w_i\mia_i(\xsus)] \\
&= \sum_{i=1}^n w_i(\expect[\mia_i(\xval)] -\expect[\mia_i(\xsus)])\\
&> 0 \text{ if } \dsus \text{ is member set}, \text{otherwise} \leq 0.
\end{aligned}
\end{equation}
Here, $w_i\in\real$ is the weight for the MIA score $\mia_i$.
Assuming that $\dsus$ and $\dval$ are i.i.d., we have $\expect[\mia_i(\xval)]>\expect[\mia_i(\xsus)]$ on member set and $\expect[\mia_i(\xval)]\approx\expect[\mia_i(\xsus)]$ on non-member set for each MIA score.
Therefore, we have $\ydiff$ close to $0$ on the non-member set, regardless of the weights $w_i$.
However, when we synthesize the \validation set with a generator, there can be a small distributional shift between $\dsus$ and $\dval$.
With this shift, we can have $\expect[\mia_i(\xval)] < \expect[\mia_i(\xsus)]$ on non-member set.
This is often the case for generated text, because the generator usually produces \validation texts that are \textit{simpler} than human-written texts, therefore causing the generated \validation texts to have smaller perplexity.
This affects most perplexity-based methods, such as LOSS, Min-K\%, and Zlib ratio.
Consequently, the linear regression algorithm can assign negative weight $w_i$ to such MIA scores, which causes $\ydiff>0$ on the non-member set and therefore high false positive rates.
To ensure that this generation shift does not add up to a falsely high $\ydiff$, we constrain the weights to be positive.
As shown in \cref{fig:constraint}, the weights are projected from $\real$ to $(0,1)$ with Sigmoid function $\sigma(x)=\frac{1}{1+e^{-x}}$.
With such a weight constraint, the linear regression only assigns a small weight $w_i$ for $\mia_i$ if $\expect[\mia_i(\xval)] < \expect[\mia_i(\xsus)]$, avoiding false positives in many cases.
We also present the empirical analysis of the weight constraint in \cref{sec:ablation} and an example for the weight constraint in \cref{app:weight_example}.

\section{Experimental Evaluation}
We start by introducing our experimental setup, further detailed in \cref{app:implement}.
Then, we present the results of DI executed based on our generated \validation data. We also perform ablation studies to investigate the contribution of each component in our proposed method. Finally, we analyze the impact of t-test sample size and the classifier architecture.

\begin{table}[t!]
    \scriptsize
    \begin{center}
    \caption{\textbf{Results for single author blog posts.} Here, p-value $<$ 0.05 indicates the suspect set is member set.}
    \vspace{5pt}
    \label{table:blog_result}
    \begin{tabular}{ccccc}
        \toprule
        True & $\auctext$ & $\auccomb$ & \multirow{2}{*}{P-value} & Inferred\\
        Membership & (\%) & (\%) &  & Membership\\
        \cmidrule{1-5}\morecmidrules\cmidrule{1-5}
        \red{\cmark} & 53.8 & 55.6 & 0.01 & \red{\cmark} \\
        \green{\xmark} & 53.8 & 53.9 & 0.13 & \green{\xmark} \\ 
        \bottomrule
    \end{tabular}
    \end{center}
    \vspace{-12pt}
\end{table}

\subsection{Experimental Setup}\label{sec:exp_setting}
\textbf{Single author data.}
We collect 1400 blog posts from a single author. 
All figures, tables, videos, and hyperlinks are removed during pre-processing and only plain text is used for evaluation.
We sample 450 posts as member data and finetune a Pythia 410M deduplicated model as target model.
The other posts are held out as non-member and \validation sets for the evaluation.

\textbf{More Complicated Dataset and Model.}
We also evaluate our method on the Pile dataset \cite{gao2020pile}, which is much more complicated and has subsets of diverse types of texts.
We use the de-duplicated version of Pythia 1B model as the target model. The training split of the Pile dataset is used as member data, and the \validation and test split is used as non-member data. 
Here, we only evaluate Pile subsets that are free from copyright issues.
Please also refer to \cref{sec:pile_config} for detailed configuration on the Pile.

\textbf{Implementation Details}
We finetune a Llama 3 8B model \cite{dubey2024llama} with LoRA as the generator.
For both types of datasets, we split 2,000 sequences as the generator inference set, and the others as the generator training split.
Both text classifier and combined classifier are trained on 1,000 synthetic \validation data and 1,000 suspect data for each dataset.
Our proposed t-test is also conducted on 1,000 synthetic \validation data and 1,000 suspect data.
More implementation details can be found in \cref{app:implement}. We also provide an analysis of hyperparameter sensitivity in \cref{app:hyperparameters}.

\subsection{Results for Single Author Dataset}

The experimental results on the single author dataset are presented in \cref{table:blog_result}. 
On the member set, the combined classifier $\ccomb$  outperforms the text classifier $\ctext$, by a large margin of 1.8\% AUC score. 
Moreover, the observed p-value of 0.01 strongly supports the alternative hypothesis, indicating that the superior performance of $\ccomb$ over $\ctext$ is statistically significant. 
This enables our method to correctly identify that the target set is part of the training set. 
For the non-member set, $\ccomb$ and $\ctext$ achieve comparable AUC scores, with a p-value of 0.13 that significantly exceeds the threshold of 0.05. 
This result confirms the ability of our approach to correctly identify non-member texts as such, thus avoiding the false positives that occur with the original LLM DI approach.
Here, we finetune the target model on the single author dataset with LoRA for one epoch.
We also present the results with other fine-tuning setups in \cref{sec:finetune_blog}.

\renewcommand{\mycolspace}{1pt}
\addtolength{\tabcolsep}{-\mycolspace} 
\begin{table}[t!]
   \scriptsize
   \caption{\textbf{Results for different Pile subsets.} \textit{True} represents the true membership while \textit{Inferred} denotes the inferred membership. Our generation is successful if these two align.}
   \label{table:pile_result}
   \begin{center}
   \begin{threeparttable}
   \begin{tabular}{cccccc}
       \toprule
       \multirow{2}{*}{Subset} & \multirow{2}{*}{True} & $\auctext$ & $\auccomb$ & \multirow{2}{*}{P-value} & \multirow{2}{*}{Inferred} \\
       &  & (\%) &  (\%) &  &  \\
       \cmidrule{1-6}\morecmidrules\cmidrule{1-6}
       \multirow{2}{*}{Pile-CC} & \red{\cmark} & 55.6 & 58.2 & 0.002 & \red{\cmark} \\
       & \green{\xmark} & 55.3 & 53.3 & 0.99 & \green{\xmark} \\
       \cmidrule{1-6}
       \multirow{2}{*}{Wikipedia} & \red{\cmark} & 56.0 & 56.8 & 0.04 & \red{\cmark} \\
        & \green{\xmark} & 54.9 & 52.4 & 1.00 & \green{\xmark} \\
       \cmidrule{1-6}
       \multirow{2}{*}{ArXiv} & \red{\cmark} & 53.6 & 59.1 & $<$0.001 & \red{\cmark} \\
       & \green{\xmark} & 53.1 & 53.3 & 0.74 & \green{\xmark} \\
       \cmidrule{1-6}
       \multirow{2}{*}{NIH ExPorter} & \red{\cmark} & 56.8 & 57.7 & 0.02 & \red{\cmark} \\
        & \green{\xmark} & 55.6 & 53.3 & 1.00 & \green{\xmark} \\
       \cmidrule{1-6}
       \multirow{2}{*}{FreeLaw} & \red{\cmark} & 52.8 & 58.4 & $<$0.001 & \red{\cmark} \\
       & \green{\xmark} & 51.4 & 53.9 & 0.09 & \green{\xmark}  \\
       \cmidrule{1-6}
       \multirow{2}{*}{Ubuntu IRC} & \red{\cmark} & 53.4 & 55.8 & 0.01 & \red{\cmark} \\
        & \green{\xmark} & 53.4 & 54.9 & 0.33 & \green{\xmark} \\
       \cmidrule{1-6}
       \multirow{2}{*}{PubMed Central} & \red{\cmark} & 54.6 & 58.1 & $<$0.001 & \red{\cmark} \\
        & \green{\xmark} & 54.7 & 55.5 & 0.11 & \green{\xmark} \\
       \cmidrule{1-6}
       \multirow{2}{*}{Github} & \red{\cmark} & 53.6 & 55.7 & 0.003 & \red{\cmark}  \\
       & \green{\xmark} & 53.9 & 55.4 & 0.07 & \green{\xmark} \\
       \cmidrule{1-6}
       \multirow{2}{*}{EuroParl} & \red{\cmark} & 51.0 & 57.0 & $<$0.001 & \red{\cmark} \\
        & \green{\xmark} & 51.4 & 53.9 & 0.07 & \green{\xmark} \\
       \cmidrule{1-6}
       \multirow{2}{*}{PhilPapers} & \red{\cmark} & 55.1 & 59.1 & $<$0.001 & \red{\cmark} \\
        & \green{\xmark} & 61.1 & 56.0 & 0.99 & \green{\xmark} \\
       \cmidrule{1-6}
       \multirow{2}{*}{HackerNews} & \red{\cmark} & 56.7 & 60.7 & $<$0.001 & \red{\cmark} \\
        & \green{\xmark} & 58.0 & 56.7 & 0.43 & \green{\xmark} \\
       \cmidrule{1-6}
       \multirow{2}{*}{Enron Emails} & \red{\cmark} & 54.5 & 58.0 & $<$0.001 & \red{\cmark} \\
        & \green{\xmark} & 54.6 & 52.9 & 0.99 & \green{\xmark} \\
       \cmidrule{1-6}
       \multirow{2}{*}{StackExchange} & \red{\cmark} & 54.3 & 60.1 & $<$0.001 & \red{\cmark} \\
        & \green{\xmark}  & 53.0 & 55.0 & 0.06 & \green{\xmark} \\
       \cmidrule{1-6}
       \multirow{2}{*}{PubMed Abstracts} & \red{\cmark} & 54.5 & 59.0 & $<$0.001 & \red{\cmark} \\
        & \green{\xmark} & 53.8 & 53.4 & 0.90 & \green{\xmark} \\
       \cmidrule{1-6}
       \multirow{2}{*}{USPTO Backgrounds} & \red{\cmark} & 52.8 & 57.1 & 0.001 & \red{\cmark} \\
        & \green{\xmark} & 52.8 & 52.2 & 0.97 &  \green{\xmark} \\
       \cmidrule{1-6}
       \multirow{2}{*}{DM Mathematics} & \red{\cmark} & 53.9 & 55.5 & 0.002 & \red{\cmark} \\
        & \green{\xmark} & 54.0 & 51.3 & 1.00 & \green{\xmark} \\
       \bottomrule
   \end{tabular}
   \end{threeparttable}
   \end{center}
   \vspace{-12pt}
\end{table}
\addtolength{\tabcolsep}{\mycolspace}

\subsection{Results for Pile Datasets}\label{sec:pythia410}
The results of different Pile subsets are shown in \cref{table:pile_result}.
We observe that DI correctly predicts the membership of datasets from diverse domains and styles, including plain text, academic writing, and code using our method for generating the \validation data.
The results also show that our generation method generalizes well to documents with different lengths, ranging from 1 KB (Wikipedia) to 70 KB (PhilPapers).
Moreover, our proposed method generalizes well to texts from different domains and languages, \eg medical (PubMed Central), legal (FreeLaw), and multilingual (EuroParl) domains.
Notably, the p-values for our difference comparison t-test are significantly lower than 0.05 on all the evaluated member sets, and higher than 0.1 on all the non-member sets.
Please refer to~\Cref{app:pythia_size} and~\Cref{app:olmo} for the results on different model sizes and model architectures.

\renewcommand{\mycolspace}{1.2pt}
\addtolength{\tabcolsep}{-\mycolspace} 
\begin{table}[t!]
    \scriptsize
    \begin{center}
    \caption{\textbf{Ablation studies of our approach.} \textit{\textbf{Setting 1-3:}} replacing our generation method with baselines. \textit{\textbf{Setting 4-5:}} removing key designs from our generation method. \textit{\textbf{Setting 6:}} without post-hoc calibration. \textit{\textbf{Setting 7:}} without weight constraint. \textit{\textbf{Setting 8:}} our complete method.}
    \vspace{4pt}
    \label{table:ablation}
\begin{tabular}{ccccc}
        \toprule
        \multirow{2}{*}{Setting}& \multirow{2}{*}{Configuration} & True & \multirow{2}{*}{P-value} & Inferred\\
        & & Membership &  & Membership\\
        \cmidrule{1-5}\morecmidrules\cmidrule{1-5}
        \multirow{2}{*}{1} & w/o Suffix Completion & \red{\cmark} & 1.0 & \green{\xmark} \\
        & \textit{(ICL Paraphrasing)}& \green{\xmark} & 1.0 & \green{\xmark} \\
        \cmidrule{1-5}
        \multirow{2}{*}{2} & w/o Suffix Completion & \red{\cmark} & 1.0 & \green{\xmark} \\
        & \textit{(ICL Text Completion)}& \green{\xmark} & 1.0 & \green{\xmark} \\
        \cmidrule{1-5}
        \multirow{2}{*}{3} & w/o Suffix Completion & \red{\cmark} & 1.0 & \green{\xmark} \\
        & \textit{(Preference Optimization)}& \green{\xmark} & 1.0 & \green{\xmark} \\
        \cmidrule{1-5}
        \multirow{2}{*}{4} & \multirow{2}{*}{w/o Segment and Shuffle} & \red{\cmark} & 1.0 & \green{\xmark} \\
        & & \green{\xmark} & 1.0 & \green{\xmark} \\
        \cmidrule{1-5}
        \multirow{2}{*}{5} & \multirow{2}{*}{w/o Suffix Comparison} & \red{\cmark} & 1.0 & \green{\xmark} \\
        & & \green{\xmark} & 1.0 & \green{\xmark} \\
        \cmidrule{1-5}
        \multirow{2}{*}{6} & w/o Post-hoc Calibration & \red{\cmark} & $<$0.001 & \red{\cmark} \\
        & \textit{(Original T-test in DI)} & \green{\xmark} & $<$0.001 & \red{\cmark} \\
        \cmidrule{1-5}
        \multirow{2}{*}{7} & \multirow{2}{*}{w/o Weight Constraint}  & \red{\cmark} & 0.004 & \red{\cmark} \\
        & & \green{\xmark} & 0.43 & \green{\xmark} \\
        \cmidrule{1-5}
        \multirow{2}{*}{8} & \multirow{2}{*}{Ours} & \red{\cmark} & $<$0.001 &  \red{\cmark}\\
        & & \green{\xmark} & 1.0 & \green{\xmark} \\
        \bottomrule
    \end{tabular}
    \end{center}
    \vspace{-12pt}
\end{table}
\addtolength{\tabcolsep}{\mycolspace}

\subsection{Ablation on Post-hoc Dataset Inference} 
\label{sec:ablation}
We conduct ablation studies to separately analyze the contribution of the three components in our \validation data generation: suffix completion, calibrating, and weight constraint.

\textbf{Suffix Completion.}
As presented in \cref{table:gen_result}, our proposed sequence completion scheme can synthesize \validation texts with a distribution much more similar to the suspect texts when compared with the baseline methods.
In addition to the AUC results, we also show that the baseline generation methods cannot produce reliable held-out sets even when combined with our post-hoc calibration and weight constraint in \cref{table:ablation}. In particular, we replace our generation scheme with three baselines, including ICL paraphrasing, ICL text completion, and preference optimization. The p-values are presented as Setting 1-3. We also remove two key designs in our generation method, 1) Segment and Shuffle, and 2) Suffix Comparison, as shown in Setting 4-5. 
In all the above settings, the p-values for both member and non-member sets are 1.0, which indicates that the $\ctext$ has better or similar performance when compared with $\ccomb$.
The reason behind the observation is that the distributional shift caused by the generation is much larger than the shift induced by the membership signal, such that $\ccomb$ does not outperform $\ctext$ even with extra membership inputs on the member set.
Consequently, the DI predicts both sets as non-member and suffers from false negatives.

\textbf{Post-hoc Calibration.}
We replace our calibration method with the original DI without calibration, as shown in Setting 6 of \cref{table:ablation}.
Specifically, only a linear classifier is optimized to aggregate different MIA metrics and output the final prediction score.
Furthermore, the t-test is conducted directly between the predictions on the target set and the ones on the \validation set.
We observe that the p-values under this condition are extremely low for both member and non-member sets, and DI has false positive in this case.
This observation aligns with results in \cref{sec:fail_case}, where we show that even a small distributional shift causes a significantly small p-value in the original DI.
Therefore, our post-hoc calibration approach is crucial to evaluating the distributional shift caused only by membership signals.

\textbf{Weight constraint.}
As explained in \cref{sec:constraint}, the weight constraint avoids summing the distributional shift caused by generation to the final MIA prediction when the direction of the generation shift is different from that caused by the membership signal.
As shown in Setting 7 of \cref{table:ablation}, applying the constraint leads to a much lower p-value on the member set and much higher on non-member set,
which helps our method make a more accurate prediction about the membership.

\begin{figure}[t] 
	\centering	\includegraphics[width=0.45\textwidth]{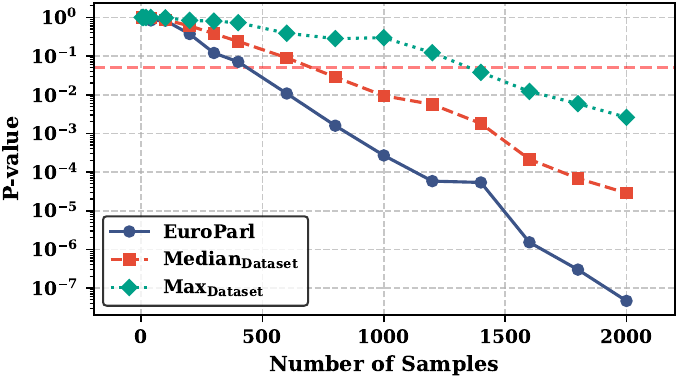}
    \vspace{-7pt}
	\caption{\textbf{The p-values of member sets with change in sample size.} $\text{Median}_\text{Dataset}$ denotes the median p-value of different datasets, and $\text{Mean}_\text{Dataset}$ is the maximum p-value of all subsets.
    Number of samples refers to the total size of both suspect and validation sets.
    }
    \vspace{-12pt}
    \label{fig:sample_ablation}
\end{figure}

\subsection{Analysis of Sample Size}\label{app:sample_size}
We also set out to analyze how the sample size in our proposed t-test affects the statistical confidence of DI with our generated \validation data.
Here, the sample size is the total number of the suspect and \validation set, which is also the number of queries made to the target model.
The two sets are of the same size, as they are produced in a pairwise manner.
We observe from \cref{fig:sample_ablation} that, as the number of samples increases, DI exhibits improved detection capability of training data. 
Notably, with fewer than 1,000 samples, DI achieves statistical significance ($p < 0.05$) across most of the evaluated datasets. When increasing the sample size to 2k queries, the method demonstrates even stronger statistical significance ($p < 0.01$) consistently across all datasets.

\begin{table}[t]
    \scriptsize
    \begin{center}
    \caption{\textbf{The AUC of different classifier architectures.}}
    \vspace{5pt}
    \label{table:classifier}
    \begin{tabular}{ccc}
        \toprule
        
        Architecture & $\auctext$ & Training Time \\
        \cmidrule{1-3}\morecmidrules\cmidrule{1-3}
        all-MiniLM-L6-v2 & 50.8 & \textbf{0.3} \\
        BERT & 51.2 & 2.0 \\
Llama3-8B, Pre-trained+LoRA & 53.2 & 65.1 \\
GPT2, Pre-trained+LoRA & 53.0 & 26.2 \\
GPT2, Pre-trained+Full Finetuned & 52.3 & 36.8 \\
        GPT2, 2 Layers+Initialized & \textbf{53.3} & 0.5 \\
        \bottomrule
    \end{tabular}
    \end{center}
    \vspace{-16pt}
\end{table}

\subsection{Choice of Classifier}

We explore different text classifier architectures and present the results for different architectures and different parameter sizes in \cref{table:classifier}. The results show that the simple GPT2-based classifier with 2 layers and initialized weights can achieve the best AUC in our experimental settings. Additionally, this lightweight classifier has a significantly short training time, making the method more practical when faced with more queries.
Therefore, we choose this 2-layer GPT2-based classifier as our text classifier. 
During our experiments, we consider the scenario where the author can only provide a limited number of tokens, so stronger text classifiers, such as Llama and full GPT2 models, can be easily overfitted.
In real-world applications, an arbitrator is suggested to select the most suitable text classifier based on their specific conditions regarding data size, data type, and computation resources.

\section{Conclusions}
We propose how to \textit{synthetically generate} an in-distribution \validation dataset to enable the real-world application of DI.
Therefore, we solve two critical challenges, namely (1) creating high-quality, diverse synthetic data that accurately reflects the original distribution and (2) bridging likelihood gaps between real and synthetic data.
Our solution relies on designing a data generator training scheme based on a suffix-based completion task and post-hoc calibration to align the likelihood gaps between real and synthetic data.
Through extensive experimental evaluation, we highlight that our method enables a robust DI and correctly identifies training data while achieving a low false positive rate. This shows our method's reliability to support copyright owners to make legitimate claims on data usage for real-world litigations.

\section*{Impact Statement}

Crawling of data for training LLMs is becoming pervasive, with model training companies scraping vast spans of the internet in order to find high-quality data. Given the strong correlation between data quality and model performance, many content creators want to protect their work from being trained on, without their consent. Claiming that one's data has been trained on, only with access to the suspect LLM, has thus far stayed near impossible. Our work takes a leap forward by allowing content creators to `post-hoc' infer if their data on the internet was trained on by leveraging synthetic data. This means that authors, bloggers, and columnists with decades of internet data can now leverage our method in order to claim their rightful ownership. Our work aims to serve as an important tool in future copyright litigations, in particular in their `discovery' period. 

\section*{Acknowledgment}
This work was supported by the German Research Foundation (DFG) within the framework of the Weave Programme under the project titled "Protecting Creativity: On the Way to Safe Generative Models" with number 545047250. Responsibility for the content of this publication lies with the authors.

\bibliography{main}

\begin{thebibliography}{46}
\providecommand{\natexlab}[1]{#1}
\providecommand{\url}[1]{\texttt{#1}}
\expandafter\ifx\csname urlstyle\endcsname\relax
  \providecommand{\doi}[1]{doi: #1}\else
  \providecommand{\doi}{doi: \begingroup \urlstyle{rm}\Url}\fi

\bibitem[Gry(2023)]{Grynbaum2023LawsuiteOpenAINewYorkTimes}
The times sues openai and microsoft over a.i. use of copyrighted work https://www.nytimes.com/2023/12/27/business/media/new-york-times-open-ai-microsoft-lawsuit.html.
\newblock 2023.
\newblock URL \url{https://www.nytimes.com/2023/12/27/business/media/new-york-times-open-ai-microsoft-lawsuit.html}.

\bibitem[Sil(2023)]{SilvermanLawsuitOpenAI}
Sarah silverman and authors sue openai and meta over copyright infringement.
\newblock 2023.
\newblock URL \url{https://www.nytimes.com/2023/07/10/arts/sarah-silverman-lawsuit-openai-meta.html}.

\bibitem[Balloccu et~al.(2024)Balloccu, Schmidtov{\'a}, Lango, and Du{\v{s}}ek]{balloccu2024leak}
Balloccu, S., Schmidtov{\'a}, P., Lango, M., and Du{\v{s}}ek, O.
\newblock Leak, cheat, repeat: Data contamination and evaluation malpractices in closed-source llms.
\newblock In \emph{Proceedings of the 18th Conference of the European Chapter of the Association for Computational Linguistics (Volume 1: Long Papers)}, pp.\  67--93, 2024.

\bibitem[Biderman et~al.(2023)Biderman, Schoelkopf, Anthony, Bradley, O’Brien, Hallahan, Khan, Purohit, Prashanth, Raff, et~al.]{biderman2023pythia}
Biderman, S., Schoelkopf, H., Anthony, Q.~G., Bradley, H., O’Brien, K., Hallahan, E., Khan, M.~A., Purohit, S., Prashanth, U.~S., Raff, E., et~al.
\newblock Pythia: A suite for analyzing large language models across training and scaling.
\newblock In \emph{International Conference on Machine Learning}, pp.\  2397--2430. PMLR, 2023.

\bibitem[Carlini et~al.(2021)Carlini, Tramer, Wallace, Jagielski, Herbert-Voss, Lee, Roberts, Brown, Song, Erlingsson, et~al.]{carlini2021extracting}
Carlini, N., Tramer, F., Wallace, E., Jagielski, M., Herbert-Voss, A., Lee, K., Roberts, A., Brown, T., Song, D., Erlingsson, U., et~al.
\newblock Extracting training data from large language models.
\newblock In \emph{30th USENIX Security Symposium (USENIX Security 21)}, pp.\  2633--2650, 2021.

\bibitem[Das et~al.(2024)Das, Zhang, and Tram{\`e}r]{das2024blind}
Das, D., Zhang, J., and Tram{\`e}r, F.
\newblock Blind baselines beat membership inference attacks for foundation models.
\newblock \emph{arXiv preprint arXiv:2406.16201}, 2024.

\bibitem[Duan et~al.(2023{\natexlab{a}})Duan, Dziedzic, Papernot, and Boenisch]{duan2023flocks}
Duan, H., Dziedzic, A., Papernot, N., and Boenisch, F.
\newblock Flocks of stochastic parrots: Differentially private prompt learning for large language models.
\newblock In \emph{Thirty-seventh Conference on Neural Information Processing Systems (NeurIPS)}, 2023{\natexlab{a}}.

\bibitem[Duan et~al.(2023{\natexlab{b}})Duan, Dziedzic, Yaghini, Papernot, and Boenisch]{duan2023privacyICL}
Duan, H., Dziedzic, A., Yaghini, M., Papernot, N., and Boenisch, F.
\newblock On the privacy risk of in-context learning.
\newblock In \emph{The 61st Annual Meeting Of The Association For Computational Linguistics}, 2023{\natexlab{b}}.

\bibitem[Duan et~al.(2024)Duan, Suri, Mireshghallah, Min, Shi, Zettlemoyer, Tsvetkov, Choi, Evans, and Hajishirzi]{duan2024membership}
Duan, M., Suri, A., Mireshghallah, N., Min, S., Shi, W., Zettlemoyer, L., Tsvetkov, Y., Choi, Y., Evans, D., and Hajishirzi, H.
\newblock Do membership inference attacks work on large language models?
\newblock \emph{arXiv preprint arXiv:2402.07841}, 2024.

\bibitem[Duarte et~al.(2024)Duarte, Zhao, Oliveira, and Li]{duarte2024cop}
Duarte, A.~V., Zhao, X., Oliveira, A.~L., and Li, L.
\newblock De-cop: Detecting copyrighted content in language models training data.
\newblock \emph{arXiv preprint arXiv:2402.09910}, 2024.

\bibitem[Dubey et~al.(2024)Dubey, Jauhri, Pandey, Kadian, Al-Dahle, Letman, Mathur, Schelten, Yang, Fan, et~al.]{dubey2024llama}
Dubey, A., Jauhri, A., Pandey, A., Kadian, A., Al-Dahle, A., Letman, A., Mathur, A., Schelten, A., Yang, A., Fan, A., et~al.
\newblock The llama 3 herd of models.
\newblock \emph{arXiv preprint arXiv:2407.21783}, 2024.

\bibitem[Dubiński et~al.(2025)Dubiński, Kowalczuk, Boenisch, and Dziedzic]{dubinski2024cdi}
Dubiński, J., Kowalczuk, A., Boenisch, F., and Dziedzic, A.
\newblock {CDI: Copyrighted Data Identification in Diffusion Models}.
\newblock In \emph{The IEEE CVF Computer Vision and Pattern Recognition Conference (CVPR)}, 2025.

\bibitem[Dziedzic et~al.(2022{\natexlab{a}})Dziedzic, Dhawan, Kaleem, Guan, and Papernot]{sslextractions2022icml}
Dziedzic, A., Dhawan, N., Kaleem, M.~A., Guan, J., and Papernot, N.
\newblock On the difficulty of defending self-supervised learning against model extraction.
\newblock In \emph{ICML (International Conference on Machine Learning)}, 2022{\natexlab{a}}.

\bibitem[Dziedzic et~al.(2022{\natexlab{b}})Dziedzic, Duan, Kaleem, Dhawan, Guan, Cattan, Boenisch, and Papernot]{dziedzic2022dataset}
Dziedzic, A., Duan, H., Kaleem, M.~A., Dhawan, N., Guan, J., Cattan, Y., Boenisch, F., and Papernot, N.
\newblock Dataset inference for self-supervised models.
\newblock \emph{Advances in Neural Information Processing Systems}, 35:\penalty0 12058--12070, 2022{\natexlab{b}}.

\bibitem[Fu et~al.(2024)Fu, Wang, Gao, Liu, Li, and Jiang]{fu2024membership}
Fu, W., Wang, H., Gao, C., Liu, G., Li, Y., and Jiang, T.
\newblock Membership inference attacks against fine-tuned large language models via self-prompt calibration.
\newblock In \emph{The Thirty-eighth Annual Conference on Neural Information Processing Systems}, 2024.

\bibitem[Gao et~al.(2020)Gao, Biderman, Black, Golding, Hoppe, Foster, Phang, He, Thite, Nabeshima, et~al.]{gao2020pile}
Gao, L., Biderman, S., Black, S., Golding, L., Hoppe, T., Foster, C., Phang, J., He, H., Thite, A., Nabeshima, N., et~al.
\newblock The pile: An 800gb dataset of diverse text for language modeling.
\newblock \emph{arXiv preprint arXiv:2101.00027}, 2020.

\bibitem[Golchin \& Surdeanu(2023)Golchin and Surdeanu]{golchin2023time}
Golchin, S. and Surdeanu, M.
\newblock Time travel in llms: Tracing data contamination in large language models.
\newblock \emph{arXiv preprint arXiv:2308.08493}, 2023.

\bibitem[Groeneveld et~al.(2024)Groeneveld, Beltagy, Walsh, Bhagia, Kinney, Tafjord, Jha, Ivison, Magnusson, Wang, et~al.]{groeneveld2024olmo}
Groeneveld, D., Beltagy, I., Walsh, E., Bhagia, A., Kinney, R., Tafjord, O., Jha, A., Ivison, H., Magnusson, I., Wang, Y., et~al.
\newblock Olmo: Accelerating the science of language models.
\newblock In \emph{Proceedings of the 62nd Annual Meeting of the Association for Computational Linguistics (Volume 1: Long Papers)}, pp.\  15789--15809, 2024.

\bibitem[Hanke et~al.(2024)Hanke, Blanchard, Boenisch, Olatunji, Backes, and Dziedzic]{hanke2024openLLMs}
Hanke, V., Blanchard, T., Boenisch, F., Olatunji, I.~E., Backes, M., and Dziedzic, A.
\newblock Open llms are necessary for current private adaptations and outperform their closed alternatives.
\newblock In \emph{Thirty-Eighth Conference on Neural Information Processing Systems (NeurIPS)}, 2024.

\bibitem[Hayes et~al.(2025)Hayes, Shumailov, Choquette-Choo, Jagielski, Kaissis, Lee, Nasr, Ghalebikesabi, Mireshghallah, Annamalai, Shilov, Meeus, de~Montjoye, Boenisch, Dziedzic, and Cooper]{hayes2025strong}
Hayes, J., Shumailov, I., Choquette-Choo, C.~A., Jagielski, M., Kaissis, G., Lee, K., Nasr, M., Ghalebikesabi, S., Mireshghallah, N., Annamalai, M. S. M.~S., Shilov, I., Meeus, M., de~Montjoye, Y.-A., Boenisch, F., Dziedzic, A., and Cooper, A.~F.
\newblock Strong membership inference attacks on massive datasets and (moderately) large language models.
\newblock 2025.

\bibitem[Kazmi et~al.(2024)Kazmi, Lautraite, Akbari, Tang, Soroco, Wang, Gambs, and L{\'e}cuyer]{kazmi2024panoramia}
Kazmi, M., Lautraite, H., Akbari, A., Tang, Q., Soroco, M., Wang, T., Gambs, S., and L{\'e}cuyer, M.
\newblock {PANORAMIA}: Privacy auditing of machine learning models without retraining.
\newblock In \emph{The Thirty-eighth Annual Conference on Neural Information Processing Systems}, 2024.

\bibitem[Kowalczuk et~al.(2025)Kowalczuk, Dubiński, Boenisch, and Dziedzic]{kowalczuk2025privacy}
Kowalczuk, A., Dubiński, J., Boenisch, F., and Dziedzic, A.
\newblock Privacy attacks on image autoregressive models.
\newblock In \emph{Forty-Second International Conference on Machine Learning (ICML)}, 2025.

\bibitem[Li \& Flanigan(2024)Li and Flanigan]{li2024task}
Li, C. and Flanigan, J.
\newblock Task contamination: Language models may not be few-shot anymore.
\newblock In \emph{Proceedings of the AAAI Conference on Artificial Intelligence}, volume~38, pp.\  18471--18480, 2024.

\bibitem[Magnusson et~al.(2024)Magnusson, Bhagia, Hofmann, Soldaini, Jha, Tafjord, Schwenk, Walsh, Elazar, Lo, et~al.]{magnusson2024paloma}
Magnusson, I., Bhagia, A., Hofmann, V., Soldaini, L., Jha, A.~H., Tafjord, O., Schwenk, D., Walsh, E., Elazar, Y., Lo, K., et~al.
\newblock Paloma: A benchmark for evaluating language model fit.
\newblock \emph{Advances in Neural Information Processing Systems}, 37:\penalty0 64338--64376, 2024.

\bibitem[Maini \& Suri(2024)Maini and Suri]{maini2024reassessingemnlp2024s}
Maini, P. and Suri, A.
\newblock Reassessing emnlp 2024’s best paper: Does divergence-based calibration for membership inference attacks hold up?
\newblock 2024.
\newblock URL \url{https://www.anshumansuri.com/blog/2024/calibrated-mia/}.
\newblock Accessed January 29, 2025.

\bibitem[Maini et~al.(2021)Maini, Yaghini, and Papernot]{maini2021dataset}
Maini, P., Yaghini, M., and Papernot, N.
\newblock Dataset inference: Ownership resolution in machine learning.
\newblock In \emph{9th International Conference on Learning Representations, {ICLR} 2021, Virtual Event, Austria, May 3-7, 2021}. OpenReview.net, 2021.

\bibitem[Maini et~al.(2024)Maini, Jia, Papernot, and Dziedzic]{maini2024llm}
Maini, P., Jia, H., Papernot, N., and Dziedzic, A.
\newblock {LLM} dataset inference: Did you train on my dataset?
\newblock In \emph{The Thirty-eighth Annual Conference on Neural Information Processing Systems}, 2024.

\bibitem[Mattern et~al.(2023)Mattern, Mireshghallah, Jin, Schoelkopf, Sachan, and Berg-Kirkpatrick]{mattern2023membership}
Mattern, J., Mireshghallah, F., Jin, Z., Schoelkopf, B., Sachan, M., and Berg-Kirkpatrick, T.
\newblock Membership inference attacks against language models via neighbourhood comparison.
\newblock In \emph{Findings of the Association for Computational Linguistics: ACL 2023}, pp.\  11330--11343, 2023.

\bibitem[Meeus et~al.(2024)Meeus, Shilov, Jain, Faysse, Rei, and de~Montjoye]{meeus2024sok}
Meeus, M., Shilov, I., Jain, S., Faysse, M., Rei, M., and de~Montjoye, Y.-A.
\newblock Sok: Membership inference attacks on llms are rushing nowhere (and how to fix it).
\newblock \emph{arXiv preprint arXiv:2406.17975}, 2024.

\bibitem[Meng et~al.(2024)Meng, Xia, and Chen]{meng2024simpo}
Meng, Y., Xia, M., and Chen, D.
\newblock Sim{PO}: Simple preference optimization with a reference-free reward.
\newblock In \emph{The Thirty-eighth Annual Conference on Neural Information Processing Systems}, 2024.

\bibitem[Oren et~al.(2024)Oren, Meister, Chatterji, Ladhak, and Hashimoto]{oren2024proving}
Oren, Y., Meister, N., Chatterji, N.~S., Ladhak, F., and Hashimoto, T.
\newblock Proving test set contamination in black-box language models.
\newblock In \emph{The Twelfth International Conference on Learning Representations}, 2024.

\bibitem[Penedo et~al.(2024)Penedo, Kydl{\'\i}{\v{c}}ek, Lozhkov, Mitchell, Raffel, Von~Werra, Wolf, et~al.]{penedo2024fineweb}
Penedo, G., Kydl{\'\i}{\v{c}}ek, H., Lozhkov, A., Mitchell, M., Raffel, C., Von~Werra, L., Wolf, T., et~al.
\newblock The fineweb datasets: Decanting the web for the finest text data at scale.
\newblock \emph{arXiv preprint arXiv:2406.17557}, 2024.

\bibitem[Rafailov et~al.(2024)Rafailov, Sharma, Mitchell, Manning, Ermon, and Finn]{rafailov2024direct}
Rafailov, R., Sharma, A., Mitchell, E., Manning, C.~D., Ermon, S., and Finn, C.
\newblock Direct preference optimization: Your language model is secretly a reward model.
\newblock \emph{Advances in Neural Information Processing Systems}, 36, 2024.

\bibitem[Rahman \& Santacana(2023)Rahman and Santacana]{rahman2023BeyondFairUse}
Rahman, N. and Santacana, E.
\newblock Beyond fair use: Legal risk evaluation for training llms on copyrighted text.
\newblock 2023.
\newblock URL \url{https://genlaw.org/CameraReady/57.pdf}.

\bibitem[Reuters(2023)]{gettyimages_lawsuit}
Reuters.
\newblock Getty images lawsuit says stability ai misused photos to train \text{AI}, 2023.
\newblock URL \url{https://www.reuters.com/legal/getty-images-lawsuit-says-stability-ai-misused-photos-train-ai-2023-02-06/}.

\bibitem[Roberts et~al.(2024)Roberts, Thakur, Herlihy, White, and Dooley]{roberts2024cutoff}
Roberts, M., Thakur, H., Herlihy, C., White, C., and Dooley, S.
\newblock To the cutoff... and beyond? a longitudinal perspective on llm data contamination.
\newblock In \emph{The Twelfth International Conference on Learning Representations}, 2024.

\bibitem[Shi et~al.(2024)Shi, Ajith, Xia, Huang, Liu, Blevins, Chen, and Zettlemoyer]{shi2024detecting}
Shi, W., Ajith, A., Xia, M., Huang, Y., Liu, D., Blevins, T., Chen, D., and Zettlemoyer, L.
\newblock Detecting pretraining data from large language models.
\newblock In \emph{The Twelfth International Conference on Learning Representations}, 2024.

\bibitem[Shokri et~al.(2017)Shokri, Stronati, Song, and Shmatikov]{shokri2017membership}
Shokri, R., Stronati, M., Song, C., and Shmatikov, V.
\newblock Membership inference attacks against machine learning models.
\newblock In \emph{2017 IEEE symposium on security and privacy (SP)}, pp.\  3--18. IEEE, 2017.

\bibitem[{\v{S}}id{\'a}k(1967)]{vsidak1967rectangular}
{\v{S}}id{\'a}k, Z.
\newblock Rectangular confidence regions for the means of multivariate normal distributions.
\newblock \emph{Journal of the American statistical association}, 62\penalty0 (318):\penalty0 626--633, 1967.

\bibitem[Soldaini et~al.(2024)Soldaini, Kinney, Bhagia, Schwenk, Atkinson, Authur, Bogin, Chandu, Dumas, Elazar, et~al.]{soldaini2024dolma}
Soldaini, L., Kinney, R., Bhagia, A., Schwenk, D., Atkinson, D., Authur, R., Bogin, B., Chandu, K., Dumas, J., Elazar, Y., et~al.
\newblock Dolma: an open corpus of three trillion tokens for language model pretraining research.
\newblock In \emph{Proceedings of the 62nd Annual Meeting of the Association for Computational Linguistics (Volume 1: Long Papers)}, pp.\  15725--15788, 2024.

\bibitem[Weber et~al.(2024)Weber, Fu, Anthony, Oren, Adams, Alexandrov, Lyu, Nguyen, Yao, Adams, Athiwaratkun, Chalamala, Chen, Ryabinin, Dao, Liang, Ré, Rish, and Zhang]{weber2024redpajama}
Weber, M., Fu, D.~Y., Anthony, Q., Oren, Y., Adams, S., Alexandrov, A., Lyu, X., Nguyen, H., Yao, X., Adams, V., Athiwaratkun, B., Chalamala, R., Chen, K., Ryabinin, M., Dao, T., Liang, P., Ré, C., Rish, I., and Zhang, C.
\newblock Redpajama: an open dataset for training large language models.
\newblock \emph{NeurIPS Datasets and Benchmarks Track}, 2024.

\bibitem[Wu et~al.(2023)Wu, Duan, and Ni]{wu2023unveiling}
Wu, X., Duan, R., and Ni, J.
\newblock Unveiling security, privacy, and ethical concerns of chatgpt.
\newblock \emph{Journal of Information and Intelligence}, 2023.

\bibitem[Xu et~al.(2024)Xu, Sharaf, Chen, Tan, Shen, Durme, Murray, and Kim]{xu2024contrastive}
Xu, H., Sharaf, A., Chen, Y., Tan, W., Shen, L., Durme, B.~V., Murray, K., and Kim, Y.~J.
\newblock Contrastive preference optimization: Pushing the boundaries of {LLM} performance in machine translation.
\newblock In \emph{Forty-first International Conference on Machine Learning}, 2024.
\newblock URL \url{https://openreview.net/forum?id=51iwkioZpn}.

\bibitem[Yeom et~al.(2018)Yeom, Giacomelli, Fredrikson, and Jha]{yeom2018privacy}
Yeom, S., Giacomelli, I., Fredrikson, M., and Jha, S.
\newblock Privacy risk in machine learning: Analyzing the connection to overfitting.
\newblock In \emph{2018 IEEE 31st computer security foundations symposium (CSF)}, pp.\  268--282. IEEE, 2018.

\bibitem[Zhang et~al.(2024{\natexlab{a}})Zhang, Das, Kamath, and Tram{\`e}r]{zhang2024membership}
Zhang, J., Das, D., Kamath, G., and Tram{\`e}r, F.
\newblock Membership inference attacks cannot prove that a model was trained on your data.
\newblock \emph{arXiv preprint arXiv:2409.19798}, 2024{\natexlab{a}}.

\bibitem[Zhang et~al.(2024{\natexlab{b}})Zhang, Sun, Yeats, Ouyang, Kuo, Zhang, Yang, and Li]{zhang2024min}
Zhang, J., Sun, J., Yeats, E., Ouyang, Y., Kuo, M., Zhang, J., Yang, H.~F., and Li, H.
\newblock Min-k\%++: Improved baseline for detecting pre-training data from large language models.
\newblock \emph{arXiv preprint arXiv:2404.02936}, 2024{\natexlab{b}}.

\end{thebibliography}
\bibliographystyle{icml2025}

\newpage
\appendix
\onecolumn
\renewcommand{\tablename}{Table}
\setcounter{table}{0}
\renewcommand{\thetable}{A\arabic{table}}
\renewcommand{\figurename}{Figure}
\setcounter{figure}{0}
\renewcommand{\thefigure}{A\arabic{figure}}

\section{Details of ICL Generation}\label{app:gpt4}

We experiment with GPT-4-Turbo and prompt it using in-context-learning learning to paraphrase the suspect data.
Each prompt includes a few data points as demonstrations (shots) and requests the model to produce paraphrases for the suspect set.
We test two types of templates to prompt GPT-4-Turbo model with in-context-learning (ICL).
One is to paraphrase the given sample based on the examples, and the other is to complete the text with part of the sentence.

\subsection{Text Completion Prompt}
In the text completion prompt, we give ten examples from the author provided documents, and prompt the GPT-4-Turbo model to complete the given sample, which is shown at the end of the prompt.
Here is an example of this kind of queries:

\textit{Input:}

\begin{lstlisting}
I have some text samples. Please help me complete the last sample based on my example samples.
A device may be provided to sense an environment's present temperature. For example, [...] a thermistor that is robust and reliable and that can be fabricated at a reasonable cost.
[...]
This invention relates to a method of synthesizing musical tones.A musical tone or natural tone produced by an acoustic or natural musical instrument can [...]  With the music synthesizer it is considerably difficult to obtain dynamic audio spectra.pneumoniae.
Radio communications receivers commonly receive and decode messages which contain information. A part of this information may be [...]a single radio will transmit messages to the remainder of the group. These radio
\end{lstlisting}

\textit{Output:}
\begin{lstlisting}
messages are selectively received by the grouped radios which decode the identity information within the [...] and secure group communications in diverse and often challenging operational environments.
\end{lstlisting}

\subsection{Paraphrasing Prompt}
In the paraphrasing prompt, we give ten examples from the author provided documents, and prompt the GPT-4-Turbo model to paraphrase the last.
Here is an example of this kind of queries:

\textit{Input:}
\begin{lstlisting}
Paraphrase the last sample according to the other samples.
1. Field of the Invention This invention relates generally to fluid injectors. More particularly, this invention relates to a proportional needle control fuel injector having a [...] The needle valve is moveable between a first position closing the injection fuel outlet port and a second position to open to injection fuel outlet.
[...]
The present invention relates to the structure of a casing of a miniature portable [...] is forcibly urged away from the casing by a tool, it is likely that the casing is scratched or the base plate is deformed.
This invention relates to fuel cells and, in particular, to improvements in the catalyst used in internal  [...]  fuel cell in which the catalyst assembly has minimum displacement and permits vertical and horizontal orientations of the fuel cell.
\end{lstlisting}

\textit{Output:}
\begin{lstlisting}
This invention pertains to fuel cells, specifically addressing enhancements in catalysts utilized  [...] and efficiency of internal reforming fuel cells by mitigating previous limitations and improving overall system functionality.
\end{lstlisting}

\subsection{Distribution Shift Analysis}
The distributional shifts are quantified in BoW AUC and GPT2 AUC as shown in \cref{table:auc_gpt}.
As observed, both types of prompt templates produces synthetic texts with large distributional shifts to the suspect sets.
Notably, the GPT2-based classfier can achieve as much as an AUC of 99.2\%.

\begin{table}[!htbp]
    \scriptsize
    \caption{\textbf{Distributional shifts between the suspect set and GPT-4-Turbo generated validation set.}}
    \label{table:auc_gpt}
    \vspace{5pt}
    \begin{center}
    \begin{tabular}{ccc}
        \toprule
          Template Type & BoW AUC(\%) & GPT2 AUC(\%) \\
        \cmidrule{1-3}\morecmidrules\cmidrule{1-3}
        ICL Text Completion & 79.2 & 99.2 \\
        ICL Paraphrasing & 76.2 & 99.0 \\
        \bottomrule
    \end{tabular}
    \end{center}
\end{table}

\section{Details of Preference Optimization Generation} \label{sec:pref_opt}
Preference optimization methods focus on optimizing a pre-trained LLM based on human preference \citep{rafailov2024direct,xu2024contrastive}.
Particularly, LLMs iteratively produce random generations, then human annotators are requested to label the generations as chosen or rejected, and the LLMs are further optimized according to this human feedback.
We note that, we can leverage preference optimization approaches to make our generator model prefer the human-written texts over synthetic data, thus producing texts with a more similar distribution to natural texts.
Here, we instantiate the preference optimization scheme with a state-of-the-art method, the simple preference optimization (SimPO) \cite{meng2024simpo}.
During each training iteration, the human-written suspect data are always labeled as chosen and the generations from the last iteration are marked as rejected.
As noted in \cref{sec:gen_method}, this approach improves significantly upon prompted paraphrasing, but still causes a large distributional shift between the suspect set and the generated \validation set.

\begin{table}[!htbp]
    \scriptsize
    \begin{center}
    \caption{\textbf{Segmentation configurations for different Pile subsets.}}
    \vspace{5pt}
    \label{table:pile_config}
    \begin{tabular}{ccccc}
        \toprule
        \multirow{2}{*}{Subset} & Number of & Chosen & Max. Snippets & Number of \\
         & Test Set in Pile &Split Size & per Document & Tokens per Snippet \\
        \cmidrule{1-5}\morecmidrules\cmidrule{1-5}
        Pile-CC & $>$3000 & 3000 & 30 & 64\\
        StackExchange & $>$4000 & 4000 & 5 & 64\\
        PubMed Abstracts & $>$6000 & 6000 & 5 & 64 \\
        Wikipedia (en) & $>$3000 & 3000 & 30 & 64\\
        USPTO Backgrounds & $>$6000 & 6000 & 10 & 32\\
        PubMed Central & $>$500 & 500 & 100 & 64\\
        FreeLaw & $>$4000 & 4000 & 5 & 32\\
        ArXiv & $>$200 & 200 & 100 & 32\\
        NIH ExPorter & $>$4000 & 4000 & 5 & 64\\
        HackerNews & $>$3000 & 3000 & 10 & 64\\
        Github  & $>$2000 & 2000 & 20 & 64\\
        Enron Emails & 1957 & 1957 & 50 & 32\\
        DM Mathematics & $>$1500 & 1500 & 50 & 32\\
        EuroParl & 290 & 290 & 200 & 32 \\
        PhilPapers & 132 & 132 & 100 & 32 \\
        Ubuntu IRC & 43 & 20 & 2000 & 32 \\
        \bottomrule
    \end{tabular}
    \end{center}
    
    \label{tab:pile_seg}
\end{table}

\section{Pile Dataset Segmentation} \label{sec:pile_config}
We present the details for the configurations of Pile subset in \cref{table:pile_config}.
We note that, it is claimed that the following Pile subsets may have copyright issues and cannot be included for evaluation: Books3, OpenWebText2, Gutenberg (PG-19), OpenSubtitles. BookCorpus2, and YoutubeSubtitles.
For most subset there are documents that are much longer than the other documents, which causes that too many snippets are sample from these documents if all snippets are used.
Therefore, we set a maximum snippet for each document on each subset according to the median lengths of the documents.
Also, we note that our approach can achieve good performance on most subsets with only 32 tokens.
For certain subsets, we use a token length of 64 for a stronger membership signal.
The average token number for the Pile subset is 45.
In practical applications, we suggest choosing the token numbers when the text-classifier has the minimal AUC to avoid distributional shift as much as possible.
As a more conservative approach, this avoids making false accusations of copyright violation.

\section{Implementation Details} \label{app:implement}
\subsection{Generator}
The LoRA rank for the generator is 32. The generator is trained for 100 epochs, and the learning rate is set to $2\times 10^{-4}$. 
We set a warm-up ratio of 0.03, and a linear scheduler is used to dynamically adjust the learning rate.

\subsection{Text and Combined Classifiers}
For both the text and the combined classifier, we leverage the basic architecture of the GPT2 classfier with an extra linear layer.
Specifically, the classifier has only two layers, with an embedding dimension of 1600 and an attention head number of 25.
As explained in \cref{sec:constraint},\label{app:gen_example}, we apply a weight constraint to the linear layer.
The GPT2-based classifier is optimized for 20 epochs, and the linear layer is further optimized for 200 epochs.

\section{Examples of Synthetic Texts}\label{app:example}
In this section, we provide some examples of the synthetic texts on the Pile dataset.
Here, prefix denotes the first half of the generated text, real suffix refers to the original suffix of the natural text, and generated suffix refers to the synthetic completion based on the prefix. 
We observe that, the generated suffixes are reasonable continuation of the prefixes.
The generated suffixes also align with the style of each dataset and do not overfit to the content of the real suffixes.

\subsection{Pile-CC}
\textit{Prefix:}
\begin{lstlisting}
are excited about and also what we hoped to see from this years E3!
\end{lstlisting}
\textit{Real suffix:}
\begin{lstlisting}
From the surprising new Spider-Man PS4 game to the bizarre We Happy Few and
\end{lstlisting}
\textit{Generated suffix:}
\begin{lstlisting}
Let us know your thoughts on this monologue as we are preparing for our next
\end{lstlisting}

\subsection{StackExchange}

\textit{Prefix:}
\begin{lstlisting}
var FKEntityListWithCastCopy = new debiteur().GetType().GetProperty(\""
\end{lstlisting}
\textit{Real suffix:}
\begin{lstlisting}
schakeling\").GetValue(dbEntry) as List<FKEntity>;//Just
\end{lstlisting}
\textit{Generated suffix:}
\begin{lstlisting}
FKEntityList\").GetValue(instance, null);\n            foreach(var t in FKEntity
\end{lstlisting}

\subsection{PubMed Abstracts}
\textit{Prefix:}
\begin{lstlisting}
were calculated using the Kaplan-Meier method. Of the 117 patients in
\end{lstlisting}
\textit{Real suffix:}
\begin{lstlisting}
whom data were analyzed, 103 had follow-up MR or CT images and 14 patients were
\end{lstlisting}
\textit{Generated suffix:}
\begin{lstlisting}
the study (76 with UC and 41 with DC), 45 patients required proctocolic resection
\end{lstlisting}

\subsection{Wikipedia (en)}
\textit{Prefix:}
\begin{lstlisting}
Em is going away for a while. While it's not up to the standard
\end{lstlisting}
\textit{Real suffix:}
\begin{lstlisting}
of "Mockingbird," it is more fully realized than the two other new
\end{lstlisting}
\textit{Generated suffix:}
\begin{lstlisting}
of their three previous albums, cattle call is still an enjoyable romp, 
\end{lstlisting}

\subsection{USPTO Backgrounds}
\textit{Prefix:}
\begin{lstlisting}
1. Field of the Invention\nThis invention relates to a storage device for athletic equipment and, in particular, to a portable storage device for transporting and retaining

\end{lstlisting}
\textit{Real suffix:}
\begin{lstlisting}
elongate items of athletic equipment such as hockey sticks and related athletic equipment.\n2. Discussion of Related Art\nNumerous team athletic activities require individual players on the
\end{lstlisting}
\textit{Generated suffix:}
\begin{lstlisting}
multiple pairs of basketballs.\n2. Description of the Related Art\nDuring the summer and other periods when there is an extended break from an athletic school or program
\end{lstlisting}

\subsection{PubMed Central}
\textit{Prefix:}
\begin{lstlisting}
example, both cycles apply Lewis acidic metal centers to bind the monomers (ep
\end{lstlisting}
\textit{Real suffix:}
\begin{lstlisting}
oxide or lactone), and both invoke labile metal alkoxide intermediates as
\end{lstlisting}
\textit{Generated suffix:}
\begin{lstlisting}
oxides or cyclic carbonates), but the axes of the metallacycle in
\end{lstlisting}

\subsection{FreeLaw}
\textit{Prefix:}
\begin{lstlisting}
Court, 638 P.2d 65 (Colo.1981
\end{lstlisting}
\textit{Real suffix:}
\begin{lstlisting}
Here, the juvenile court denied the GAL's motions because it did not want
\end{lstlisting}
\textit{Generated suffix:}
\begin{lstlisting}
), cert. denied, 454 U.S. 1146, 102
\end{lstlisting}

\subsection{Arxiv}
\textit{Prefix:}
\begin{lstlisting}
up and vice versa. In contrast, fundamentalists expect the price to track its
\end{lstlisting}
\textit{Real suffix:}
\begin{lstlisting}
fundamental value. Orders from this type of agent may be written as\n\n$$D
\end{lstlisting}
\textit{Generated suffix:}
\begin{lstlisting}
underlying fundamentals up and down, but given sufficient acceleration the price might \u201crun away
\end{lstlisting}

\subsection{NIH ExPorter}
\textit{Prefix:}
\begin{lstlisting}
attachment and growth, respectively. Together with an industrial sponsor, Vaxiron,
\end{lstlisting}
\textit{Real suffix:}
\begin{lstlisting}
Inc., we will develop quality control tools and metrics for assessing vaccine antigen formulations,
\end{lstlisting}
\textit{Generated suffix:}
\begin{lstlisting}
the applicant has carried out clinical trials of different vaccine candidates based on different viruses for
\end{lstlisting}

\subsection{Github}
\textit{Prefix:}
\begin{lstlisting}
.string \"reach only by using a BIKE technique.$\"\n\nRoute110_Text_
\end{lstlisting}
\textit{Real suffix:}
\begin{lstlisting}
16EEF6:: @ 816EEF6\n\t.string \"Which
\end{lstlisting}
\textit{Generated suffix:}
\begin{lstlisting}
16F381:: @ 816F381\n\t.string \"ROUTE {ROAD
\end{lstlisting}

\subsection{Enron Emails}
\textit{Prefix:}
\begin{lstlisting}
Lay.  He went on to say that Kenneth was Dewayne Re
\end{lstlisting}
\textit{Real suffix:}
\begin{lstlisting}
es' cousin  and started telling about all of your fine attributes and what a
\end{lstlisting}
\textit{Generated suffix:}
\begin{lstlisting}
ams' direct \nreport and that it would be extremely difficult for Kenneth to get
\end{lstlisting}

\subsection{EuroParl}
\textit{Prefix:}
\begin{lstlisting}
het mondeling amendement op schrift heeft gekregen.\nIk st
\end{lstlisting}
\textit{Real suffix:}
\begin{lstlisting}
el voor om niet te spreken over \"de Raad en de lidstat
\end{lstlisting}
\textit{Generated suffix:}
\begin{lstlisting}
akk voor de uitnodiging om tijdens uw volgende bij
\end{lstlisting}

\subsection{PhilPapers}
\textit{Prefix:}
\begin{lstlisting}
distribute well among [the gods who fought with him] their titles and privileges
\end{lstlisting}
\textit{Real suffix:}
\begin{lstlisting}
" (885, cf. 66\u201367 and 74); to swallow
\end{lstlisting}
\textit{Generated suffix:}
\begin{lstlisting}
 (17.1). Orderly distribution of praise for the victory is re
\end{lstlisting}

\subsection{Ubuntu IRC}
\textit{Prefix:}
\begin{lstlisting}
about setting up reoccuring status meetings?\n<dfarning> should we start
\end{lstlisting}
\textit{Real suffix:}
\begin{lstlisting}
holding those or is it too soon?\n<dfarning> Luke will be joining
\end{lstlisting}
\textit{Generated suffix:}
\begin{lstlisting}
with a status meeting or a design meeting?\n<manusheel> dfarning
\end{lstlisting}

\subsection{HackerNews}
\textit{Prefix:}
\begin{lstlisting}
Angular (work just uses Dojo).\n\nPeople don't seem to
\end{lstlisting}
\textit{Real suffix:}
\begin{lstlisting}
be hungry here.\n\n------\nlewispollard\nWorked for IBM as a software engineer on one of 
\end{lstlisting}
\textit{Generated suffix:}
\begin{lstlisting}
care that it's adding yet another ~20KB per page. We're\nsaying no 
\end{lstlisting}

\section{Generalization to Different Model Sizes}\label{app:pythia_size}

Here, we evaluate the performance of our method on three different model sizes of the Pythia model: 1.4B, 2.8B, and 6.9B. We use an outlier removal ratio $r=0.01$ for Pythia-6.9B, $r=0.05$ for Pythia-2.8B, $r=0.1$ for Pythia-1.4B, $r=0.15$ for Pythia-1B, $r=0.2$ for Pythia-410M. 
All the tested models are the deduplicated versions.
The results demonstrate that our proposed method generalizes well across different model sizes.

\renewcommand{\mycolspace}{2pt}
\newcommand{\smallerspace}{5.3pt}
\addtolength{\tabcolsep}{-\mycolspace} 
\begin{table}[h]
   \tiny
   \caption{\textbf{Results for different sizes of Pythia Models.} \textit{True} represents the true membership while \textit{Inferred} denotes the inferred membership. Our generation is successful if these two align.}
   \label{table:model_size}
    \vspace{5pt}
   \begin{center}
   \begin{threeparttable}
   \begin{tabular}{c@{\hspace{5pt}}c@{\hspace{10pt}}c@{\hspace{\smallerspace}}c@{\hspace{\smallerspace}}c@{\hspace{\smallerspace}}c@{\hspace{\smallerspace}}c@{\hspace{\smallerspace}}c@{\hspace{\smallerspace}}c@{\hspace{\smallerspace}}c@{\hspace{\smallerspace}}c@{\hspace{\smallerspace}}c@{\hspace{\smallerspace}}c@{\hspace{\smallerspace}}c@{\hspace{\smallerspace}}c@{\hspace{\smallerspace}}c@{\hspace{\smallerspace}}c@{\hspace{\smallerspace}}c}
       \toprule
       \multirow{3}{*}{Subset} & \multirow{3}{*}{True} & \multicolumn{4}{c}{Pythia-410M} & \multicolumn{4}{c}{Pythia-1.4B} & \multicolumn{4}{c}{Pythia-2.8B} & \multicolumn{4}{c}{Pythia-6.9B} \\
       \cmidrule{3-18}
       & & $\auctext$ & $\auccomb$ & \multirow{2}{*}{P-value} & \multirow{2}{*}{Inferred} & $\auctext$ & $\auccomb$ & \multirow{2}{*}{P-value} & \multirow{2}{*}{Inferred} & $\auctext$ & $\auccomb$ & \multirow{2}{*}{P-value} & \multirow{2}{*}{Inferred} & $\auctext$ & $\auccomb$ & \multirow{2}{*}{P-value} & \multirow{2}{*}{Inferred} \\
       &  & (\%) &  (\%) &  &  & (\%) &  (\%) &  &  & (\%) &  (\%) &  &  & (\%) &  (\%) &  & \\
       \cmidrule{1-18}\morecmidrules\cmidrule{1-18}
       \multirow{2}{*}{Pile-CC} & \red{\cmark} & 54.9 & 57.4 & 0.009 & \red{\cmark} & 55.5 & 57.5 & 0.007 & \red{\cmark} & 56.2 & 57.8 & 0.02 & \red{\cmark} & 55.9 & 58.2 & 0.006 & \red{\cmark} \\
       & \green{\xmark} & 54.2 & 51.0 & 1.00 & \green{\xmark} & 55.1 & 53.9 & 0.89 & \green{\xmark} & 54.0 & 52.8 & 0.94 & \green{\xmark} & 54.8 & 55.1 & 0.30 & \green{\xmark} \\
       \cmidrule{1-18}
       \multirow{2}{*}{ArXiv} & \red{\cmark} & 54.3 & 60.5 & $<$0.001 &  \red{\cmark} & 54.0 & 59.3 & $<$0.001 &  \red{\cmark} & 53.8 & 57.7 & $<$0.001 &  \red{\cmark}  & 53.7 & 57.0 & $<$0.001 & \red{\cmark} \\
       & \green{\xmark} & 52.8 & 52.9 & 0.87 &  \green{\xmark} & 53.8 & 54.0 & 0.64 &  \green{\xmark} & 52.3 & 53.1 & 0.67 &  \green{\xmark}  & 52.4 & 53.6 & 0.39 & \green{\xmark} \\
       \cmidrule{1-18}
       \multirow{2}{*}{FreeLaw} & \red{\cmark}  & 52.9 & 58.3 & $<$0.001 &  \red{\cmark} & 52.6 & 57.3 & $<$0.001 &  \red{\cmark} & 52.2 & 57.4 & $<$0.001 &  \red{\cmark} & 52.2 & 56.7 & $<$0.001 & \red{\cmark} \\
       & \green{\xmark}  & 52.3 & 53.6 & 0.41 &  \green{\xmark} & 52.1 & 54.7 & 0.07 &  \green{\xmark} & 52.2 & 54.6 & 0.05 &  \green{\xmark} & 51.9 & 54.6 & 0.05 & \green{\xmark} \\
       \cmidrule{1-18}
       PubMed & \red{\cmark}  & 54.7 & 58.0 & $<$0.001 &  \red{\cmark} & 54.4 & 58.4 & $<$0.001 &  \red{\cmark} & 55.1 & 57.9 & 0.002 &  \red{\cmark} & 54.7 & 57.2 & 0.002 & \red{\cmark} \\
       Central & \green{\xmark}  & 55.2 & 55.4 & 0.24 &  \green{\xmark} & 54.9 & 55.9 & 0.06 &  \green{\xmark} & 55.1 & 56.0 & 0.06 &  \green{\xmark} & 54.9 & 55.5 & 0.10 & \green{\xmark} \\
       \cmidrule{1-18}
       Euro- & \red{\cmark}  & 51.2 & 55.7 & 0.002 &  \red{\cmark} & 51.5 & 55.6 & 0.004 &  \red{\cmark} & 51.0 & 53.9 & 0.02 &  \red{\cmark} & 50.7 & 54.2 & 0.04 & \red{\cmark} \\
       Parl & \green{\xmark}  & 51.3 & 53.8 & 0.13 &  \green{\xmark} & 51.3 & 53.2 & 0.31 &  \green{\xmark} & 51.1 & 53.2 & 0.20 &  \green{\xmark} & 51.4 & 53.3 & 0.22 & \green{\xmark} \\
       \cmidrule{1-18}
       Phil- & \red{\cmark}  & 55.3 & 59.5 & $<$0.001 &  \red{\cmark} & 54.8 & 58.3 & $<$0.001 &  \red{\cmark} & 54.9 & 58.0 & $<$0.001 &  \red{\cmark} & 55.2 & 56.8 & 0.02 & \red{\cmark} \\
       Papers & \green{\xmark}  & 61.2 & 55.6 & 1.00 &  \green{\xmark} & 61.2 & 56.1 & 0.98 &  \green{\xmark} & 61.4 & 56.2 & 0.98 &  \green{\xmark} & 60.2 & 56.0 & 0.99 & \green{\xmark} \\
       \cmidrule{1-18}
       Hacker & \red{\cmark}  & 56.4 & 61.1 & $<$0.001 &  \red{\cmark} & 56.5 & 59.8 & $<$0.001 &  \red{\cmark} & 56.3 & 59.0 & $<$0.001 &  \red{\cmark} & 56.5 & 59.1 & 0.002 & \red{\cmark} \\
       News & \green{\xmark}  & 58.2 & 55.6 & 0.94 &  \green{\xmark} & 58.2 & 57.1 & 0.24 &  \green{\xmark} & 58.1 & 57.7 & 0.17 &  \green{\xmark} & 58.7 & 57.7 & 0.24 & \green{\xmark} \\
       \cmidrule{1-18}
       Enron & \red{\cmark}  & 54.5 & 58.1 & $<$0.001 &  \red{\cmark} & 54.4 & 56.9 & 0.003 &  \red{\cmark} & 54.1 & 57.7 & $<$0.001 &  \red{\cmark} & 54.4 & 56.2 & 0.04 & \red{\cmark} \\
       Emails & \green{\xmark}  & 54.8 & 52.8 & 1.00 &  \green{\xmark} & 54.6 & 53.0 & 0.97 &  \green{\xmark} & 54.5 & 54.9 & 0.20 &  \green{\xmark} & 54.6 & 54.0 & 0.76 & \green{\xmark} \\
       \cmidrule{1-18}
       Stack & \red{\cmark}  & 54.1 & 61.9 & $<$0.001 &  \red{\cmark} & 54.2 & 58.8 & $<$0.001 &  \red{\cmark} & 53.9 & 58.1 & $<$0.001 &  \red{\cmark} & 54.2 & 57.2 & 0.002 & \red{\cmark} \\
       Exchange & \green{\xmark}   & 52.3 & 55.0 & 0.06 &  \green{\xmark} & 52.1 & 54.3 & 0.090 &  \green{\xmark} & 52.1 & 54.1 & 0.24 &  \green{\xmark} & 52.5 & 54.0 & 0.29 & \green{\xmark} \\
       \cmidrule{1-18}
       PubMed & \red{\cmark}  & 54.6 & 59.7 & $<$0.001 &  \red{\cmark} & 54.4 & 58.5 & $<$0.001 &  \red{\cmark} & 54.1 & 58.3 & $<$0.001 &  \red{\cmark} & 54.4 & 58.1 & $<$0.001 & \red{\cmark} \\
       Abstract & \green{\xmark}  & 54.9 & 52.6 & 1.00 &  \green{\xmark} & 54.1 & 54.0 & 0.74 &  \green{\xmark} & 53.9 & 53.7 & 0.83 &  \green{\xmark} & 53.9 & 54.3 & 0.57 & \green{\xmark} \\
       \cmidrule{1-18}
       USPTO & \red{\cmark}  & 52.9 & 56.8 & 0.002 & \red{\cmark} & 52.8 & 56.3 & 0.004 & \red{\cmark} & 52.4 & 55.6 & 0.028 & \red{\cmark} & 52.6 & 55.4 & 0.018 & \red{\cmark} \\
       Back. & \green{\xmark}  & 52.6 & 51.7 & 1.00 &  \green{\xmark} & 52.9 & 52.2 & 0.99 &  \green{\xmark} & 52.5 & 52.2 & 0.98 & \green{\xmark} & 52.6 & 53.1 & 0.75 & \green{\xmark} \\
       \bottomrule
   \end{tabular}
   \end{threeparttable}
   \end{center}
   \vspace{-12pt}
\end{table}
\addtolength{\tabcolsep}{\mycolspace}

\section{Example of Weight Constraint}\label{app:weight_example}
Here, we provide the following example to illustrate the importance of weight constraint.
In \cref{tab:weight_example}, we show the score of three MIAs for a suspect/held-out pair on both member and non-member sets.
For each suspect/held-out pair, the smaller \text{MIA} score is highlighted in \textbf{bold} in the table. We have the following observations:

\begin{enumerate}
    \item On the \textit{member} set, suspect data consistently shows smaller \text{MIA} scores. This occurs because membership signals have stronger effects than generation, causing suspect data to consistently yield lower \text{MIA} scores than held-out data.
    
    \item On the \textit{non-member} set, held-out data may exhibit smaller values for certain MIAs. This happens because generation randomness introduces fluctuation in \text{MIA} scores.
\end{enumerate}

For both member and non-member sets, we train a linear model $l$ that aggregates all \text{MIA} scores to predict an overall score:
\begin{equation}
l(x) = \sum_i w_i \text{MIA}_i(x)
\end{equation}

The held-out set is labeled as 1 and the suspect set as 0. The model assigns positive weights $w_i$ to any \text{MIA} metrics $\text{MIA}_i$ on the \textit{member} set because label $0 < 1$ and $\text{MIA}_i$(suspect) $< \text{MIA}_i$(held-out). However, on the \textit{non-member} set, the model assigns a negative weight $w_3$ for $\text{MIA}_3$. This means a smaller $\text{MIA}_3$ score in the held-out set would contribute to a larger overall \text{MIA} score, which is undesirable. To address this, we constrain all weights in the linear model to be strictly positive, ensuring that a lower $\text{MIA}_i$ score can only result in a lower overall \text{MIA} score.

\begin{table}[h]
   \scriptsize
   \caption{\textbf{An example to demonstrate the importance of the weight constraint.}}
   \label{tab:weight_example}
   \vspace{5pt}
   \begin{center}
   \begin{tabular}{ccccccc}
       \toprule
       Membership & Split & Label & $\text{MIA}_1$ & $\text{MIA}_2$ & $\text{MIA}_3$ \\
       \cmidrule{1-6}\morecmidrules\cmidrule{1-6}
       \multirow{2}{*}{\red{\cmark}} & Suspect \textit{(natural)} & 0 & \textbf{0.86} & \textbf{0.87} & \textbf{0.54} \\
       & Held-out \textit{(generated)} & 1 & 0.90 & 0.91 & 0.55 \\
       \cmidrule{1-6}
       \multirow{2}{*}{\green{\xmark}} & Suspect \textit{(natural)} & 0 & \textbf{0.88} & \textbf{0.89} & 0.58 \\
       & Held-out \textit{(generated)} & 1 & 0.90 & 0.90 & \textbf{0.56} \\
       \bottomrule
   \end{tabular}
   \end{center}
   \vspace{-12pt}
\end{table}

\section{Other Finetuning Configurations for Single Author Dataset} \label{sec:finetune_blog}
We evaluate our proposed approach on the single author dataset under different finetuning settings in \cref{tab:finetune_blog}.
The results show that, the membership signal is stronger when the model is fine-tuned with more epochs.
Also, our method performs better when full finetuning is used instead of LoRA.

\begin{table}[h]
   \scriptsize
   \caption{\textbf{Results for different fine-tuning methods. }\textit{True} represents the true membership while \textit{Inferred} denotes the inferred membership. Our generation is successful if these two align.}
   \label{tab:finetune_blog}
   \vspace{5pt}
   \begin{center}
   \begin{tabular}{cccccc}
       \toprule
       \multirow{2}{*}{Fine-tuning Method} & \multirow{2}{*}{True} & $\auctext$ & $\auccomb$ & \multirow{2}{*}{P-value} & \multirow{2}{*}{Inferred} \\
       &  & (\%) &  (\%) & &  \\
       \cmidrule{1-6}\morecmidrules\cmidrule{1-6}
       LoRA  & \red{\cmark} & 53.8 & 55.6 & 0.01 & \red{\cmark} \\
       (1 epoch) & \green{\xmark} & 53.8 & 53.9 & 0.13 & \green{\xmark} \\
       \cmidrule{1-6}
       LoRA & \red{\cmark} & 53.7 & 56.2 & 0.005 & \red{\cmark} \\
       (10 epochs) & \green{\xmark} & 53.6 & 53.5 & 0.14 & \green{\xmark} \\
       \cmidrule{1-6}
       Full Finetuning & \red{\cmark} & 53.7 & 56.8 & 0.008 & \red{\cmark} \\
       (1 epoch) & \green{\xmark} & 53.8 & 53.7 & 0.21 & \green{\xmark} \\
       \bottomrule
   \end{tabular}
   \end{center}
   \vspace{-12pt}
\end{table}

\section{Results on the OLMo Model}\label{app:olmo}
We conduct the experiments to analyze the performance with OLMo-7B model \citep{groeneveld2024olmo}.
The OLMo-7B model is trained on the Dolma V.1.7 dataset \citep{soldaini2024dolma}, which has a large size of 4.5 TB. 
Following \citet{duan2024membership}, we use Dolma V.1.7 as the member set and employ Paloma \citep{magnusson2024paloma} as the non-member set. 
The results in \cref{table:olmo_result} demonstrate that our method successfully detects both member and non-member sets for Wikipedia and Common Crawl subsets when using the OLMo-7B model as the target model.

\begin{table}[h]
   \scriptsize
   \caption{\textbf{Results for OLMo-7B on different data subsets.} \textit{True} represents the true membership while \textit{Inferred} denotes the inferred membership. Our generation is successful if these two align.}
   \label{table:olmo_result}
   \vspace{5pt}
   \begin{center}
   \begin{tabular}{cccccc}
       \toprule
       Subset & True & $\auctext$ (\%) & $\auccomb$ (\%) & P-value & Inferred \\
       \cmidrule{1-6}\morecmidrules\cmidrule{1-6}
       \multirow{2}{*}{Wikipedia} & \red{\cmark} & 52.9 & 55.4 & 0.009 & \red{\cmark} \\
        & \green{\xmark} & 52.1 & 50.6 & 1.0 & \green{\xmark} \\
       \cmidrule{1-6}
       \multirow{2}{*}{Common Crawl} & \red{\cmark} & 53.5 & 55.7 & 0.01 & \red{\cmark} \\
        & \green{\xmark} & 54.2 & 53.8 & 0.68 & \green{\xmark} \\
       \bottomrule
   \end{tabular}
   \end{center}
   \vspace{-12pt}
\end{table}

\section{Ablation Studies on Single Author Dataset}
In addition to the ablation studies on the Pile presented in \cref{sec:ablation}, we also perform the ablation studies on the single author dataset. The results in \cref{tab:ablation_blog} follow a similar trend to the Pile, showing the importance of each component in our framework.

\begin{table}[h]
   \scriptsize
   \caption{\textbf{Results for different configurations.} \textit{True Membership} represents the true membership while \textit{Inferred Membership} denotes the inferred membership. Our generation is successful if these two align.}\label{tab:ablation_blog}
   \vspace{5pt}
   \begin{center}
   \begin{tabular}{cccc}
       \toprule
       Configuration & True Membership & P-value & Inferred Membership \\
       \cmidrule{1-4}\morecmidrules\cmidrule{1-4}
       w/o Suffix Completion & \red{\cmark} & 1.0 & \green{\xmark} \\
       (ICL Paraphrasing) & \green{\xmark} & 1.0 & \green{\xmark} \\
       \cmidrule{1-4}
       w/o Post-hoc Calibration & \red{\cmark} & $<$0.001 & \red{\cmark} \\
       (Original T-test in DI) & \green{\xmark} & $<$0.001 & \red{\cmark} \\
       \cmidrule{1-4}
       w/o Weight & \red{\cmark} & 0.02 & \red{\cmark} \\
       Constraint & \green{\xmark} & 0.08 & \green{\xmark} \\
       \cmidrule{1-4}
       \multirow{2}{*}{Ours} & \red{\cmark} & 0.01 & \red{\cmark} \\
       & \green{\xmark} & 0.13 & \green{\xmark} \\
       \bottomrule
   \end{tabular}
   \end{center}
   \vspace{-12pt}
\end{table}

\section{Analysis of Hyperparameter Sensitivity}\label{app:hyperparameters}

We conducted a comprehensive analysis of hyperparameter sensitivity, focusing on two key parameters: the number of epochs and the number of t-test samples. 
The number of epochs represents the training epochs for our linear model that aggregates MIA scores. 
The number of t-test samples indicates the total sample size used in our statistical analysis, including both the suspect and synthetic held-out sets. 
Our experimental results in \cref{table:hyperparameters} demonstrate that our proposed method exhibits robust performance across a wide range of values for both hyperparameters, indicating low sensitivity to these configuration choices.

\begin{table}[h]
    \scriptsize
    \begin{center}
    \caption{\textbf{Performance of our method across different numbers of epochs and T-test samples.}}
    \vspace{5pt}
    \label{table:hyperparameters}
    \begin{tabular}{ccccc}
        \toprule
        
        Hyperparameter & Value & True membership & P-value & Inferred membership \\
        \cmidrule{1-5}\morecmidrules\cmidrule{1-5}
        \multirow{8}{*}{Number of Epochs} & \multirow{2}{*}{100} & \red{\cmark} & $<$0.001 & \red{\cmark} \\
        & & \green{\xmark} & 1.0 & \green{\xmark} \\
        \cmidrule{2-5}
        & \multirow{2}{*}{200} & \red{\cmark} & $<$0.001 & \red{\cmark} \\
        & & \green{\xmark} & 1.0 & \green{\xmark} \\
        \cmidrule{2-5}
        & \multirow{2}{*}{500} & \red{\cmark} & $<$0.001 & \red{\cmark} \\
        & & \green{\xmark} & 1.0 & \green{\xmark} \\
        \cmidrule{2-5}
        & \multirow{2}{*}{1000} & \red{\cmark} & 0.003 & \red{\cmark} \\
        & & \green{\xmark} & 1.0 & \green{\xmark} \\
        \midrule
        \multirow{8}{*}{Number of T-test Samples} & \multirow{2}{*}{1000} & \red{\cmark} & $<$0.001 & \red{\cmark} \\
        & & \green{\xmark} & 1.0 & \green{\xmark} \\
        \cmidrule{2-5}
        & \multirow{2}{*}{2000} & \red{\cmark} & $<$0.001 & \red{\cmark} \\
        & & \green{\xmark} & 1.0 & \green{\xmark} \\
        \cmidrule{2-5}
        & \multirow{2}{*}{3000} & \red{\cmark} & $<$0.001 & \red{\cmark} \\
        & & \green{\xmark} & 0.41 & \green{\xmark} \\
        \cmidrule{2-5}
        & \multirow{2}{*}{4000} & \red{\cmark} & $<$0.001 & \red{\cmark} \\
        & & \green{\xmark} & 0.25 & \green{\xmark} \\
        \bottomrule
    \end{tabular}
    \end{center}
    \vspace{-12pt}
\end{table}

\section{Other Related Works about Test Set Contamination Detection}
Test set contamination is a newly identified risk, where the public test benchmarks are involved during LLM training \citep{balloccu2024leak}.
For example, \citet{roberts2024cutoff} observe that LLMs are better at generating code with more appearances on GitHub, revealing that LLMs can be contaminated with open-source GitHub data and are overestimated on coding tasks.
Similarly, \citet{li2024task} demonstrate that some LLMs have a better performance on few-shot benchmarks constructed before the model training, which indicates test set contamination for LLMs.
To detect test set contamination, \citet{golchin2023time} design prompts that guide LLM to reproduce exact or near-exact test set instances, such that the model encloses the contaminated samples memorized during the pre-training phase.
\citet{oren2024proving} compare the target model predictions between a test set and all of its permutations.
However, this method is based on the assumption that the test set is involved in the training set in its exact order, which could be interrupted by a random shuffle before training.
Test set contamination can also be a potential application of our method, as the proposed approach can perform training data detection on complex datasets composed by different authors.

\section{Algorithm of Our Work}

We present the detailed algorithms for our held-out data generation in \Cref{alg:generation}, and post-hoc calibration in \Cref{alg:calibration}.

\begin{algorithm}
\caption{Held-out Data Generation}
\begin{algorithmic}[1]\label{alg:generation}
\REQUIRE  Documents $Doc = \{Doc_1, ..., Doc_m\}$
\REQUIRE  Hyperparameters: Document number $m$, Maximum sequence in each document $MaxSeq$ 
\ENSURE  Suspect set $\dsus$ and \validation set $\dval$ are nearly IID
\STATE Initialize: $Seq,\dsus,\dval = \{\},\{\},\{\}$  
\FOR{each document $doc_i\in Doc$ } 
    \STATE Segment $doc_i$ into multiple sequences $\{seq_i^1, ..., seq_i^{m_i}\}$ 
    \IF{$m_i < MaxSeq$}
        \STATE $Seq_i = \{seq_i^1, ..., seq_i^{m_i}\}$
    \ELSE
        \STATE $Seq_i =$ randomly sampled $MaxSeq$ sequences from $\{seq_i^1, ..., seq_i^{m_i}\}$
    \ENDIF
    \STATE $Seq = Seq \cup Seq_i$
\ENDFOR
\STATE Randomly split $Seq$ into generator training set $Seq_{train}$ and generator inference set $Seq_{test}$
\STATE Optimize generator $g$ on $Seq_{train}$ with next-token prediction loss
\FOR{each $seq_i \in Seq$}
    \STATE $pre_i, suf_i = \text{Divide}(seq_i)$
    \STATE $suf_i' = g(pre_i)$
    \STATE $\dsus = \dsus \cup \{(suf_i,0)\}$
    \STATE $\dval = \dval \cup \{(suf_i',1)\}$
\ENDFOR

\end{algorithmic}

\end{algorithm}

\begin{algorithm}
\caption{Post-hoc Calibration}
\begin{algorithmic}[1]\label{alg:calibration}
\REQUIRE Target model $f$
\REQUIRE Suspect set $\dsus$ and \validation set $\dval$ are nearly IID.
\STATE Randomly split $\dsus$ into suspect training set $\dsustrain$ and suspect test set $\dsustest$
\STATE Randomly split $\dval$ into \validation training set $\dvaltrain$ and \validation test set $\dvaltest$
\STATE Optimize a text classifier $c_{text}(x)$ on $\dsustrain\cup \dvaltrain$
\STATE Optimize a combined classifier $c_{comb}(x,\text{MIA}(f(x)))$ on $\dsustrain\cup \dvaltrain$
\STATE $\dtextdiff=\{\}$
\STATE $\dcombdiff=\{\}$
\FOR{$\xsustest, \xvaltest \in \dsustest,\dvaltest$}
\STATE $\dcombdiff = \dcombdiff \cup \{c_{comb}(\xvaltest,\text{MIA}(f(\xvaltest)))-c_{comb}(\xsustest,\text{MIA}(f(\xsustest)))\}$
\STATE $\dtextdiff = \dtextdiff \cup \{c_{text}(\xvaltest)-c_{text}(\xsustest)\}$
\ENDFOR
\STATE Compare and $\dcombdiff$ and $\dtextdiff$ with t-test

\end{algorithmic}

\end{algorithm}

\end{document}